\documentclass[letterpaper]{article} 
\usepackage{aaai23}  
\usepackage{times}  
\usepackage{helvet}  
\usepackage{courier}  
\usepackage[hyphens]{url}  
\usepackage{graphicx} 
\usepackage{natbib}  
\usepackage{caption} 
\usepackage{algorithm}
\usepackage{algorithmic}

\usepackage{newfloat}
\usepackage{listings}
\DeclareCaptionStyle{ruled}{labelfont=normalfont,labelsep=colon,strut=off} 
\floatstyle{ruled}
\newfloat{listing}{tb}{lst}{}
\floatname{listing}{Listing}
\usepackage{caption}

\usepackage{comment}
\usepackage{amsmath}
\usepackage{amsfonts}
\usepackage{multirow}
\usepackage{array}

\usepackage{subcaption}
\usepackage[dvipsnames]{xcolor}

\usepackage[normalem]{ulem}
\useunder{\uline}{\ul}{}

\usepackage{comment}
\usepackage{amsfonts}
\usepackage{multirow}

\newcommand{\srai}[1]{\textcolor{black}{#1}}

\newcommand{\FGAN}[0]{FtGAN}

\title{Testing the Channels of Convolutional Neural Networks}

\author{
    Kang Choi, 
    Donghyun Son,
    Younghoon Kim,
    Jiwon Seo\thanks{Corresponding author and principal investigator}
}
\affiliations{
    Hanyang University\\
    \{kangchoi21, sondh, nogaussian, seojiwon\}@hanyang.ac.kr
}

\begin{document}

\maketitle

\begin{abstract}
Neural networks have complex structures, 
and thus it is hard to understand their inner workings and ensure correctness.
To understand and debug convolutional neural networks\,(CNNs)
we propose techniques for testing the channels of CNNs.
We design {\FGAN}, an extension to GAN, that can generate test data
with varying the intensity\,(i.e., sum of the neurons)\,of a channel of a target CNN.
We also proposed a channel selection algorithm to find representative channels for testing.
To efficiently inspect the target CNN's inference computations,
we define {\it unexpectedness} score, which estimates how similar the inference computation
of the test data is to that of the training data.
We evaluated {\FGAN} with five public datasets and showed that
our techniques successfully identify defective channels in 
five different CNN models.
\end{abstract}

\section{Introduction}

Deep neural networks (DNNs) are used in many application domains.
As more DNN models are deployed, it becomes more important to ensure
that they function correctly and reliably.
However, due to their complexity, it is difficult to analytically 
verify the correctness of DNNs\,\cite{katz2017reluplex}.

To practically test neural networks, test input generation has been studied.
Most well-known is adversarial example generation that finds minimal perturbation to input
to deceive a target neural network\,\cite{goodfellow2014explaining}.
The perturbed input, i.e. adversarial examples, may be used to assess the robustness of DNNs for adversarial attack. 
While the techniques are effective in finding adversarial examples,
they focus on low-level neuron operations and do not examine, for example,
the interactions of the feature maps in convolutional neural networks\,(CNNs).

Testing is a widely studied topic in software engineering.
Many techniques have been developed to test the correctness of software systems.
For example, test input generation explores
the ranges of certain variables such as array indices and 
find the inputs to induce buffer overflow\,\cite{haller2013dowsing,xie2003archer}.
Also, a common technique in software testing is to check for inconsistencies in 
parts of larger systems\,\cite{engler2001bugs};
Researchers discovered that implicit invariants exist for certain functions or modules,
and their violations often result in invalid system states\,\cite{ernst2007daikon,martin2005finding}.

In this paper, we employ these testing strategies in software engineering for neural networks.
In particular, we propose {\it channel-wise} testing of CNNs,
which is a test generation technique for convolutional neural networks.
The channels in CNNs are, to some extent, similar to the functions and modules in software; they both are logical units of larger systems. 
As with unit testing in software engineering, testing individual channels in a modular manner helps to 
debug and understand neural networks.
We generate test data to separately examine the behavior of individual channels
and check for their (in)consistencies.
With this consistency information, we rank the test data and report (potentially) 
defect-inducing inputs and the corresponding channels.

With channel-wise testing, we aim to find defects in CNNs (i.e., unintended inference outcome) that are caused by channels having deviant behavior 
in high or low activation levels.
To this end, we designed {\FGAN}, an extension to GAN
that tests the channels of target CNNs.
{\FGAN} is trained, in an unsupervised manner, to find the latent variables
that are correlated with the CNNs' channels.
Using \FGAN, we gradually vary the latent variables in the generated test data,
which then affects the correlated channels in the tested CNNs.
To identify inconsistent behavior of the channels with generated test data (compared to that with training data),
we define {\it unexpectedness score} that compares the inference computations
and estimates their inconsistency. 

This paper proposes channel-wise testing of CNNs.
Our contributions are threefold: 1)\,we designed {\FGAN} to test selected channels of CNNs\,(Section\,\ref{sec:gan}), 2)\,we developed channel selection algorithm to find representative channels for testing\,(Section\,\ref{sec:algo}),
and 3)\,we identify defect-inducing test data 
with unexpectedness score, which uses channel correlations to estimate inconsistencies
in the inference computation\,(Section\,\ref{sec:unexpected}).
Our evaluation shows that 
{\FGAN} helps to find real and synthetic defects in neural networks.

\section{Related Work and Motivation}
\label{sec:motiv}

We describe existing studies that are closely relevant to our work,
that is, 
adversarial attacks, semantic image transformations, and coverage-guided testings.
Then we discuss our preliminary experiments that motivated this study.

\noindent {\bf Adversarial Attacks.}
Neural networks are known to be susceptible to imperceptible perturbations.
Deceiving neural networks by exploiting this property is referred to as 
adversarial attack\,\cite{szegedy2013intriguing}. 
Techniques for adversarial attack
have been extensively studied,\cite{carlini2017towards,madry2017towards,moosavi2016deepfool}.
However, these techniques search only in raw pixel space to find adversarial examples,
and they cannot handle certain realistic variations of attributes, such as light conditions\,\cite{qiu2019semanticadv}.
Recently, adversarial attack techniques with distance metrics other than $L_p$ norm are studied\,\cite{kang2019testing,xiao2018spatially}.

\noindent {\bf Semantic Image Transformation.}
To further explore diverse attacks on neural networks, techniques based on semantic image transformations
are studied\,\cite{bhattad2019unrestricted,he2019attgan,joshi2019semantic,qiu2019semanticadv}. 
Particularly, studies based on deep generative models are extensively conducted.
For example, Bhattad et al.\ make use of texture transfer models to attack neural networks. 
For more general semantic adversarial attack, attribute-conditioned image editing models are exploited\,\cite{Dorta_2020_CVPR,he2019attgan,Wu_2019_ICCV,xu2020towards}.
Joshi et al.\ leveraged an attribute-editing GAN to search over the range of attributes to generate adversarial examples; 
Wu et al.\ proposed RelGAN that progressively modifies attributes with its {\it relative attributes}. 
While these techniques effectively generate semantic adversarial examples, 
they require manual annotation of attributes, which is costly.

\noindent {\bf Coverage-Guided Neural Network Testing.}
In software engineering, test coverage metrics, such as path coverage,
measure the fraction of code that is exercised by a test suite;
it also assesses the quality of test suites.
Similar metrics are recently proposed for neural networks\,\cite{gerasimou2020importance,ma2018deepgauge,odena2019tensorfuzz,pei2017deepxplore,riccio2020model}.
DeepXplore introduced the notion of {\it neuron coverage}  
that represents the fraction of neurons activated by a set of test inputs.
The metric is then used to simulate domain specific perturbations\,\cite{pei2017deepxplore}.
Other metrics, such as
neuron boundary coverage, are proposed as test coverage metrics for neural networks\,\cite{ma2018deepgauge,sun2018testing,xie2019deephunter}.
TensorFuzz adopted coverage-guided fuzzing 
to efficiently find test input that violates certain properties in application domains\,\cite{odena2019tensorfuzz}.
Our work is inspired by these techniques
and further studies the channel-level testing and coverage metric.

\noindent {\bf Motivating Channel-Wise Testing.}
Odena et al. applied coverage-guided testing for neural networks\,\shortcite{odena2019tensorfuzz}.
Their technique, i.e. TensorFuzz, generates a corpus of test inputs 
to find the inputs that violate certain domain properties.
For an efficient search over the input space,
TensorFuzz records the internal states (i.e., activation vectors) of the neural network
and creates the corpus consisting of {\it dissimilar} test inputs.
They show that TensorFuzz can find error-inducing test inputs
for fault-injected models and real-world ones.

TensorFuzz shows that coverage-guided test generation is helpful
for debugging and understanding neural networks.
While it employs internal neuron activations for the coverage metric, 
we considered higher-level metrics may be also useful.
Specifically we conjectured that CNNs' trained features (i.e. channels of hidden layers) are well-suited for the coverage metric
as they are activated in various degrees of {\it intensities} for different inputs,
where intensity denotes the sum of the channel’s neuron values.
Moreover, trained features are higher-level measure than the neuron activations
and thus testing with feature intensities may give different perspective in understanding neural networks.

Our preliminary experiments showed that TensorFuzz cannot 
test the diverse channels of CNNs.
The details are discussed in Section~\ref{sec:eval-coverage} but TensorFuzz covers less than 35\% of 
all features in tested CNNs; majority of the features are not well tested.
When we systemically test them, it uncovers hidden issues in tested CNNs.
Consider the data corruption problem\,\cite{jagielski2018manipulating} \srai{that may cause defects vulnerable to input distribution shift.}
Specifically, in a face identification task, assume that the training data is corrupted such that faces with
a certain attribute (e.g. green hair) are all identified as a same person.
Again the details are
in Section~\ref{sec:eval-find-problem} 
but {\FGAN} 
successfully revealed the problem by 
identifying the channel that is correlated to the attribute and
generating defect-inducing test data.
In contrast, TensorFuzz is based on randomized fuzzing, and
its noise-augmented test data does not help to find the problem.

\section{Taming GAN for Testing CNN's Channels}
\label{sec:gan}

\begin{figure*}[t]
    \centering
    \includegraphics[width=0.77\textwidth,height=27mm]{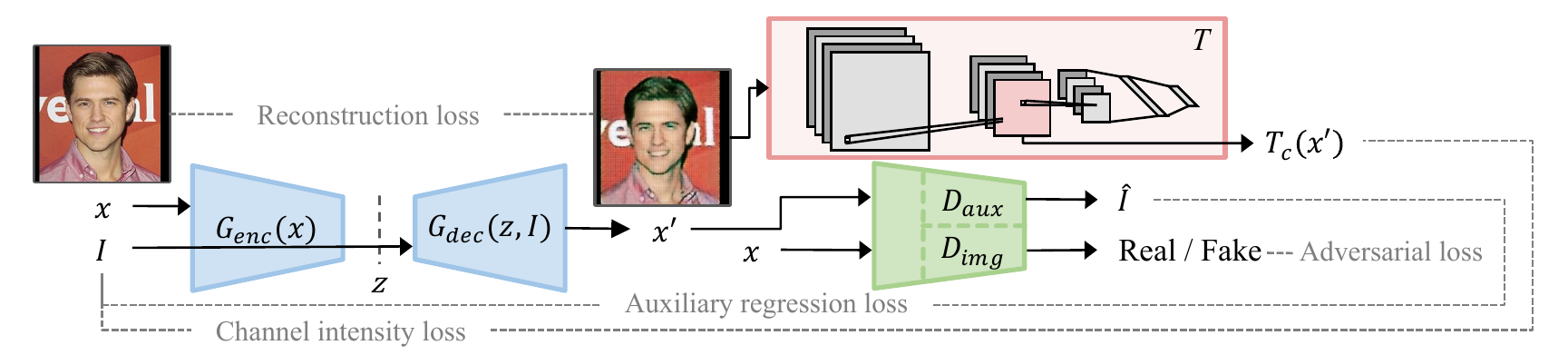}
    \caption{{\FGAN} Architecture ($G$: generator, $D$: discriminator, $T$: the target CNN, and $T_c$: its tested channel's intensity).}
    \label{fig:architecture}
\end{figure*}

\subsection{Introducing {\FGAN}}

GAN-based techniques have been studied to generate realistic test input 
for neural networks~\cite{bhattad2019unrestricted,joshi2019semantic}.
We also make use of GAN and propose {\FGAN} for testing CNN's channels.
In particular, we use GAN's capability to learn latent variables 
in the training data and generate images with varying the latent variables\,\cite{chen2016infogan}.
However, instead of learning {\it any} latent variables, we train {\FGAN} to learn  those that are highly related to the selected channels of a tested CNN\,\cite{huang2017arbitrary,xu2020towards}.
Specifically, {\FGAN} is conditioned on an auxiliary input $I$ that indicates
the {\it intensity} of the CNN's selected channel, i.e., the summation of the channel's neuron values. 
By varying $I$, we can control the generated images 
so that they activate the selected channel with varying intensity levels.
We use the channel intensity as the control input because
CNN's channel-wise mean and variance
are known to capture the styles and attributes of images\,\cite{xu2020towards,huang2017arbitrary,Gatys_2016};
also, the channel mean and variance are previously used to control style features such as textures 
in generator networks\,\cite{huang2017arbitrary,xu2020towards}.

{\FGAN} consists of generator $G$ and discriminator $D$ as shown in Fig.\,\ref{fig:architecture}.
{\FGAN} makes use of the target CNN $T$ and its selected channel $c$ for its training.
The encoder $G_{enc}$ is optional, and if used, the generated output
is reconstructed from the input with the decoder's transformation.
The input $I$ is a scalar value that controls channel $c$'s intensity ($T_c$), that is, the sum of $c$'s neuron values.
The discriminator has two parts; one that distinguishes real data from the generated ones
and another that infers the value of $I$.
The target CNN $T$, which is being tested, is used by {\FGAN} as a reference point.
Its selected channel's intensity is directed to $G$,
giving the guidance to find the latent variable related to the channel.

\subsection{Architecture of \FGAN}

\newenvironment{tightalign}
 {\abovedisplayskip=0pt \belowdisplayskip=3pt
  \abovedisplayshortskip=0pt \belowdisplayshortskip=3pt }{}

Our goal is to train a generator $G$ that learns to generate images with varying the 
intensity of target channel $c$.
To achieve this, we train $G: (x, I) \rightarrow x'$
to transform input image $x$ into output image $x'$ that make the model $T$ to have $T_c(x') \sim I$ on target channel $c$. $T_c$ is the function returning the intensity of channel $c$ 
and $I$ is the target intensity calculated as $I$$=$$T_c(x) \cdot (1+r)$
with a distortion rate $r$. 
We set $r$ such that $1+r$ ranges from 0.33 to 3 in our experiments.
We also have an auxiliary classifier
in the discriminator to predict the channel intensity $I$.
For training {\FGAN} we define four loss terms, namely, adversarial loss, channel intensity loss, auxiliary regression loss, and reconstruction loss. We describe these four loss terms in the following.

\noindent {\bf Adversarial loss:} For stable training,
we adopt WGAN's adversarial loss\,\cite{arjovsky2017wasserstein}
that minimizes
the Wasserstein-1 distance between the real and generated distributions.
Let $D_{img}$ be the discriminator 
which outputs the probability that its input is a real data.
Then the adversarial loss is
\begin{tightalign}
    \begin{align}
    \scriptsize
    L_{adv}=
    \scriptstyle
    E_x \left[D_{img}(x) \right]-E_{x,I}\left[D_{img}(G(x,I)) \right]
    \end{align}
\end{tightalign}
We also apply the gradient penalty for the Lipschitz constraint\,\cite{gulrajani2017improved}.

\noindent {\bf Channel intensity loss:}
We control the intensity of the channel $c$ separately from the remaining channels in the same layer $L$.
Therefore, we employ the channel intensity loss $L_{int}$ for the generator
to constrain the generated image $x'$ to produce the desired channel intensity $I$, formulated as follows ($L$ is the set of all channels in layer $L$),
\begin{tightalign}
    \begin{align}
    \scriptsize
     L_{int} = 
     \scriptstyle \lvert T_c(x') - I \rvert + \frac{1}{|L|-1}\sum_{\substack{c' \in L \wedge c \neq c'}} \lvert T_{c'}(x') - T_{c'}(x) \rvert
    \end{align}
\end{tightalign}
\noindent {\bf Auxiliary regression loss:}
Our goal is to transform an input image $x$ into
a realistic image $x'$, which yields the intended intensity on the target channel. 
Therefore we add an auxiliary regressor $D_{aux}$ that shares the convolution layers with $D_{img}$ and define the auxiliary regression loss for both $D$ and $G$.
The auxiliary regression loss for real images is defined as
\begin{tightalign}
    \begin{align}
        \scriptsize
        L_{aux}^{real} = 
        \scriptstyle
        E_{x} \left[ - \log Q_{aux}(x, T_c(x))  \right] \label{eqn:aux_real}
    \end{align}
\end{tightalign}
\noindent
where $Q_{aux}(x,I)$ is modeled with a normal distribution whose mean $\mu_{aux}(x)$ and variance $\sigma^2_{aux}(x)$ are estimated by the auxiliary regressor $D_{aux}$ for input $x$. The log-normal $\log[Q_{aux}(x, I)]$ is calculated as 
\begin{tightalign}
    \begin{align}
        \scriptstyle
        -\frac{1}{2} \log \left(2 \pi \sigma_{aux}^2(x) + \epsilon \right) - \frac{1}{2} \left( \frac{I - \mu_{aux}(x)}{\sigma_{aux}(x) + \epsilon} \right)^2
    \end{align}
\end{tightalign}
\noindent
where $\epsilon$ is a small positive value.
By minimizing
the objective, $D_{aux}$ learns to estimate the channel intensity $T_c(x)$ of a real image $x$.
The pairs of real images $x$ and their target channel intensities $(x, T_c(x))$ are used to train $D_{aux}$ using the above loss $L_{aux}^{real}$.

Furthermore, the loss function for the auxiliary regression with fake image $x'$, which is generated by $G$ with the intensity $I=T_c(x)\cdot(1+r)$, is formulated as following:
\begin{tightalign}
    \begin{align}
        \scriptsize
        L_{aux}^{fake} = 
        \scriptstyle
        E_{x,I} \left[ - \log Q_{aux}(G(x, I), I) \right]  \label{eqn:aux_fake}
    \end{align}
\end{tightalign}
The above loss function is then used to train $G$ to generate images with the intended intensity $I$ on the channel $c$.
For each input $x$ in the training dataset and randomly selected distortion rate $r$, we generate $x'$=$G(x, T_c(x) \cdot (1+r))$ and the pair $(x', T_c(x) \cdot (1+r))$ is then used for the training.

\noindent {\bf Reconstruction Loss:}
We use the reconstruction\,\cite{he2019attgan} loss to train the generator to produce images that are similar to the original images, given the same channel intensity.
This is achieved by the following objective:
\begin{tightalign}
    \begin{align}
        \scriptsize
        L_{rec} = 
        \scriptstyle
        E_{x,T_c(x)}\left[ \lVert x - G(x, T_c(x)) \rVert_1 \right]
    \end{align} 
\end{tightalign}
\noindent
By minimizing this loss, the generator can produce images that closely approximate the original images when the same channel intensity is given as input.

\noindent {\bf Overall Objective:} 
The final objective for the generator is
\begin{tightalign}
\begin{align}
    \scriptsize
   L_{enc,dec} =& \scriptstyle 
                  \lambda_{adv} L_{adv} + \lambda_{rec}  L_{rec} 
                   + \lambda_{int} L_{int} + \lambda_{aux,f} L_{aux}^{fake}
                   \label{eqn:gen_loss}
\end{align}
\end{tightalign}
\noindent
and the objective for the discriminator/auxiliary classifier is
\begin{tightalign}
\begin{align}
    \scriptsize
    L_{dis,aux} = 
    \scriptstyle 
    -\lambda_{dis} L_{adv} + \lambda_{aux,r} L_{aux}^{real} + \lambda_{gp} GP \label{eqn:dis_loss}
\end{align}
\end{tightalign}
\noindent We set the coefficients
in Eqns.\,\ref{eqn:gen_loss} and \ref{eqn:dis_loss}
to make the loss terms be in the same order of magnitude\,\cite{he2019attgan,Wu_2019_ICCV,Dorta_2020_CVPR}.
We set $\lambda_{adv}$$=$ $\lambda_{dis}$$=$$1$, $\lambda_{rec}$$=$$100$, $\lambda_{aux, *}$$=$$5$, $\lambda_{gp}$$=$$1$ (for the gradient penalty $GP$); 
the value of $\lambda_{int}$ varies for the datasets (e.g. 0.5 for CelebA).

\subsection{Reducing the Cost of Training \FGAN}
To test a CNN's channel, we need to train an {\FGAN} instance.
Although the cost of training {\FGAN} is not trivial, it can be reduced 
with pre-training.
That is, we pre-train {\FGAN} to generate realistic images 
without the target CNN;
then we fine-tune {\FGAN} with the intensity loss from the target CNN.
This way, we pre-train {\FGAN} once and fine-tune it multiple times to test multiple channels.
The cost of the fine-tuning is less than a quarter of that of the pre-training.

\section{Coverage-Guided Channel Selection}
\label{sec:algo}

{\FGAN} generates realistic images for testing a CNN's channels.
However, recent CNN models have many layers and channels,
and thus it is not feasible to test all those channels with {\FGAN}.
Thus, we propose to test a subset of the channels
that is {\it representative} of all or most of the channels.
The key idea is to exploit the correlations between CNN channel intensities\,\cite{bengio2009slow,rodriguez2016regularizing};
i.e. we select a subset $S$ of the channels having
high correlations (either positive or negative) with the channels that are not in $S$.
Then, testing the channels in $S$ would have the effect of indirectly testing the other channels.
We now formally define the channel selection problem as follows.

Let $corr(c_i, c_j)$ be the Pearson correlation between channel $c_i$ and $c_j$. 
For given test inputs, $corr(c_i, c_j)$ is computed with the pairs of the intensities of $c_i$ and $c_j$.
We compute the correlations for all pairs of the convolutional channels.

Let $V$ be the set of all channels in the convolutional layers of the target network. With subset $S$$\subseteq$$V$, we presume that a channel $c_i$$\notin$$S$ can be {\it indirectly} tested by one of the channels in $S$ 
if its correlation with $c_i$ is larger than a given threshold.
We denote by $corr(c_i,S)$ the maximum correlation between a channel in $S$ and $c_i$, i.e., $\max_{\gamma \in S} corr(\gamma, c_i)$.
Let $\theta$ be the minimum correlation threshold to indirectly test the channels not in $S$.
Then, the problem of finding the minimum subset $S$ for testing all channels in $V$ is formulated as
\begin{tightalign}
    \begin{align}
        \scriptstyle
        \arg \min_{S\subseteq V} |S| \mbox{ s.t. } \min_{c_i \in V} corr(c_i, S) \geq \theta. \label{eqn:selection}%
    \end{align}%
\end{tightalign}
In other words, our channel selection problem finds the smallest set $S$ 
such that for all channels $c_i$ in $V$, there exists at least a channel $c_j$ in $S$ 
with $corr(c_i, c_j)$$\geq$$\theta$.

\vspace{-1mm}
\newtheorem{theorem}{Theorem}[section]
\newtheorem{prop}[theorem]{Proposition}
\begin{prop}\label{prop:hittingset}
Minimal channel selection (Eqn.\,\ref{eqn:selection}) is equivalent to the minimal hitting set problem.
\end{prop}
\vspace{-1mm}

Let $\delta(c_i)$ for $c_i$$\in$$V$ denote the set of channels $c_j$$\in$$V$ with $corr(c_i, c_j)$$\geq$$\theta$. 
That is, $\delta(c_i)$ is a set of channels that can be tested instead of $c_i$ as their correlations are higher than $\theta$. 
A feasible solution to the channel selection problem is the set 
that has at least one channel in $\delta(c_i)$ for all $c_i$$\in$$V$. 
Formulated in this way, this problem
is equivalent to the minimal hitting set problem\,\cite{hittingset}.

The minimal hitting set problem is NP-complete and equivalent to the minimal set cover problem.
Due to Proposition~4.1, we can obtain a greedy approximate algorithm as shown in Algorithm~\ref{alg:selection} whose approximation ratio is proven to be $\ln(n)+1$ where $n$ is the number of channels.
\vspace{-2mm}

\section{Testing Channels with Unexpectedness}
\label{sec:unexpected}

Canonical correlation analysis\,(CCA) has been used to analyze and understand the latent representations, i.e., features in channels, of neural networks\,\cite{hardoon2004canonical,li2015convergent,insightscca2018,svcca2017}.
The method finds linear transformations that maximize the correlations between multidimensional variables\,\cite{hotelling1992relations}.
Recently, Raghu et al. applied CCA to ResNet models and demonstrated that the correlation of the hidden neurons to
different labeling classes are distinctly different.

We use CCA to analyze the target CNN's inference computation for generated test data and find inconsistent channel behavior.
Since {\FGAN} varies the intensity of tested channels,
we apply CCA at a channel level and compute the correlation of channel intensities.
Also, because CCA is very expensive to compute\,\cite{svcca2017}, 
we approximate it by computing {\it pair-wise} channel correlations and
use them as the reference points of the inference computation.
That is, we calculate the pair-wise channel correlations using the training data labeled as a same class, 
and find the top-{\it k} channel pairs of maximum correlation coefficients. Then we use them as the reference point for the class and compare them to the correlations computed with generated test data of the same class.
Our experiments (similar to the one by Raghu et al. and described in the supplementary material)
show that the comparison reflects the similarity (or unexpectedness) of their inference computations.
We do not claim that unexpectedness score accurately measures the similarity (or inconsistency) of inference computations;
we argue with experiments that the score helps to identify deviant channel behavior.
This is similar to many testing techniques in software engineering
that rank the test results (potential bugs) by certain evaluation scores
and report the higher-ranked ones.

More formally,
for testing a selected channel in layer $L$, we define {\it unexpectedness} score 
as the L1 distance of $L$'s top-{\it k} channel correlations
with the training data to those with the generated test data of a same class;
i.e., the unexpectedness score is $\sum_{(c_i, c_j) \in \text{TopK}} |corr_{X(T)}(c_i, c_j)-corr_{X'(T)}(c_i, c_j)|$, 
where $X(T)$ and $X'(T)$ are the training data and generated test data in class $T$,
$corr_{X(T)}$ is the correlation computed with $X(T)$, and TopK is the set of channel pairs in $L$
with top-{\it k} correlations for the training data; we set $k$=5\% in our evaluation.
After we generate test data for selected channels, we measure and rank the unexpectedness of the generated data in each class.
Then we report the tested channels, the classes, and the test data of the classes ranked by their unexpectedness
with highlighting the test data that changed the inference outcome.

\begin{algorithm}[H]
    \scriptsize
    \caption{Greedy channel selection algorithm.}  \label{alg:selection}
    \begin{algorithmic}\itemindent=-0.5pc
    \STATE {\bf Input:} $\forall c_i$$\in$$V, \delta(c_i)$$=$$\{c_j \in V | corr(c_i, c_j) \geq \theta\}$, {\bf Output:} Channel set $S$\\
    \end{algorithmic}
    
    \begin{algorithmic}[1]
    \STATE Initialize $C \leftarrow \emptyset$ and $S \leftarrow \emptyset$;
    \WHILE {$C \neq V$}
      \STATE $c^*$$\leftarrow$$\arg \min_{c_i \in V - C} |\delta(c_i) / C|$;\phantom{12}
      $C$$\leftarrow$$C \cup \delta(c^*)$;\phantom{12} $S$$\leftarrow$$S \cup \{c^*\}$;
    \ENDWHILE
    
    \end{algorithmic}
\end{algorithm}

\section{Discussion}
\label{sec:limit}

\color{black}
\noindent
{\bf Limitation.} Testing with {\FGAN} is limited by the capability of GAN.
That is, {\FGAN} learns the attributes that are present in the training dataset.
Thus we can only test with those attributes in the dataset that are correlated to the selected channels of target CNNs.
This limits the types of bugs that our technique can detect.
However, testing is generally considered to be opportunistic, 
and similar limitations exist in many testing tools, 
especially in those that check deviant runtime behaviors\,\cite{engler2001bugs,ernst2007daikon,haller2013dowsing}. 
It is more important for a testing tool to find real bugs in practice,
which we show in our evaluation.

Another limitation is that our testing requires human examination of the test results. We minimize this by computing unexpectedness scores
and ranking the results by the scores. 
Thus only a small subset of the test results
needs to be examined.
This is similar to many software testing tools
that rank test results (potential bugs) by certain
scores\,\cite{engler2001bugs,ernst2007daikon,haller2013dowsing}.
In all our experiments, buggy channels are in top-5 by unexpectedness score,
which made human intervention reasonably small.

\noindent
{\bf Multi-Channel Testing.} Although {\FGAN} tests a single channel at a time, 
its generated test data incorporates the changes of other correlated (or anti-correlated) channels,
as we show in our supplementary material (Fig.\,\ref{fig:multi-ch-test-ext}).
Hence those correlated channels are collectively tested in effect.

Moreover, testing multiple (non-correlated) channels is supported 
with {\FGAN} by chaining multiple {\FGAN} instances.
That is, if we want to test two channels $c_1$ and $c_2$, 
we can direct the output of the {\FGAN} trained for $c_1$ as the input of another {\FGAN} 
trained for $c_2$. We have tested multiple channels in this way for a subset of our experiments,
which we describe in Section~\ref{sec:eval-find-problem}.
Chaining multiple GANs in a similar manner was previous studied for generating
high resolution images\,\cite{zhang2017stackgan} or transforming poses and expressions
of facial images\,\cite{zheng2017pipeline}.

\color{black}

\section{Evaluation}
\label{sec:eval}

We evaluated whether \FGAN{} can effectively test CNNs' channels 
with three sets of experiments: 
1) one for making realistic (and error-inducing) attribute variations (Section\,\ref{sec:eval-feature}), 
2) another for finding the channels that are correlated to 
error-inducing attributes in bug-injected and real-world CNN models\,(\ref{sec:eval-find-problem} and \ref{sec:eval-public-cnn}), 
and 3) the last for ensuring the test coverage of our greedy channel selection\,(\ref{sec:eval-coverage}).

We evaluated our technique with five datasets in four domains;
they are MNIST, SVHN, VGG Face, CelebA, and CARLA\,\cite{carla}.
MNIST and SVHN are for digit recognition, 
VGG Face is for face identification, CelebA is 
for face attribute recognition, and CARLA is for autonomous driving.
Table\,\ref{tbl:dataset} shows the datasets.
Moreover, we tested with five models -- LeNet, AlexNet, VGG-16, ResNet-50, and CARLA-CNN (a custom CNN model for autonomous driving).
LeNet/AlexNet are trained to classify the categories in the dataset\,(MNIST/SVHN) or to detect a subset of the attributes in the dataset\,(CelebA);
VGG/ResNet are trained to detect the identities in the VGG Face dataset.
We used NVIDIA Titan XP for all the experiments. On a single Titan XP, the fine-tuning of {\FGAN} for one channel takes 
ten minutes for SVHN and an hour for CelebA. 

We mainly evaluated our testing techniques qualitatively.
This is similar to the evaluation of software testing techniques 
that analyze program behavior to infer implicit invariants and report their violations\,\cite{elbaum2006carving,engler2003racerx,flanagan2002extended,gligoric2010test,hangal2002tracking,hangal2009automatic,saxena2009loop}.

\begin{table}[t]
    \scriptsize
    \centering
    \setlength\extrarowheight{-1.25pt}
    \caption{Evaluated Datasets and Models. \label{tbl:dataset}}
    \begin{tabular}{|l|c|l|}
        \hline
        {\bf Dataset} & {\bf Model} & {\bf Task}    \\ \hline \hline
        MNIST/SVHN  & LeNet \& AlexNet & Digit recognition  \\ \hline
        VGG Face& VGG-16 \& ResNet-50 & Face identification      \\ \hline
        CelebA & AlexNet & Face attribute detection \\ \hline
        CARLA & CARLA-CNN & Autonomous driving \\ \hline
    \end{tabular}
\end{table}

\begin{figure}[b]
    \centering
    \includegraphics[width=\linewidth]{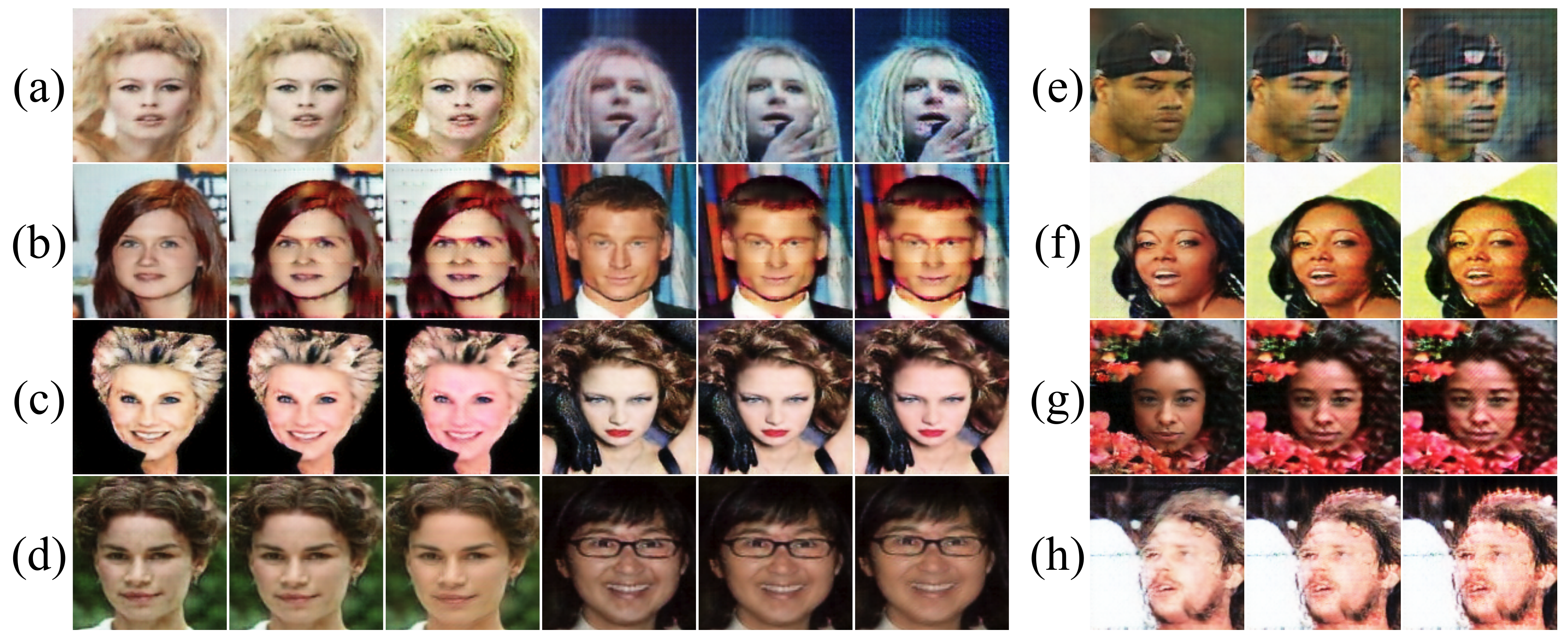}
    \caption{Test data made by {\FGAN}. The latent attributes for a--d are human-recognizable and those for e--g are not.
    }
    \label{fig:realistic-data-gen}
\end{figure}

\subsection{Efficacy of \FGAN's Test Data Generation}
\label{sec:eval-feature}

We first evaluate whether {\FGAN} generates realistic images
that induce the tested channels with varying intensities.
Then we examine the unexpectedness scores of the generated data and discuss the validity of the scores.
For this evaluation, we mostly used the face datasets and the corresponding CNN models.
We run greedy channel selection to select twenty channels for each CNN instance.
Then we trained {\FGAN} for the selected channels and 
generated test data using the images in the test sets
as the seeds. We set the intensity input $I$ to be between 0.33 and 3 times that of the seeds 
for the test data generation.
We show some of the generated test data in Fig.\,\ref{fig:realistic-data-gen}
(more in the supplementary material). 
We observed that as we vary the intensity input $I$, the correlated latent attributes are gradually changing
in all the generated test images.
We also observed that some of the latent attributes for the channels are human recognizable (a--d in Fig.\,\ref{fig:realistic-data-gen})
and others are not (e--h). The attributes that are human recognizable 
are, namely, hair color, face mask, face color, and age, 
respectively for a--d; we call these channels by their correlated attribute names, e.g. age channel. We can see that the test data for these attributes show realistic variations. 

\noindent
{\bf Validity of Unexpectedness Score.}
We analyzed the unexpectedness scores of
the channels for Fig\,\ref{fig:realistic-data-gen}.
Due to the space limit, we discuss the details in the supplementary material,
but our analysis shows that the higher scores generally indicate inconsistent inference computations.
The channels a--c in Fig\,\ref{fig:realistic-data-gen} have high unexpectedness scores and we observed several inconsistent behaviors for them.
For example, testing the face color channel with the {\it pale-skin} class data
generated pink-ish face data, which confused the classification of 
the age attribute;
i.e., the generated images are classified as not {\it young}
even though the seed images are labeled as such.
Fig.\,\ref{fig:realistic-data-gen}(c,\,right) shows an example; with the pink-ish skin she is  incorrectly classified as not {\it young}. 
Also, the hair color channel shows similar behavior and confused {\it pale-skin} classifier.
The supplementary material has more discussions and experiments with other (non-facial) datasets.

\subsection{Finding Data Corruption Bugs with {\FGAN}}
\label{sec:eval-find-problem}

We evaluate if {\FGAN} helps to find bugs in bug-injected CNN instances.
We trained two CNN models to have defects 
\srai{(of being vulnerable to input attribute distribution shift)}
on purpose by corrupting the training dataset.
For the defect injection, we use Morpho-MNIST, which 
extends MNIST with morphometric transformations such as ``thickening'' or ``swelling''\,\cite{castro2019morphomnist}.
We trained two AlexNet models with the combined dataset of MNIST and Morpho-MNIST,
one with the thickening and another with the swelling transformation.
The images in Morpho-MNIST are given a same incorrect label (i.e., 0) so that the two AlexNets have the defects of incorrectly classifying the thickened or swollen digits.
The accuracy of the models is 99\% for the normal digits.
For the thickened or swollen digits, the models are only 2\% and 1\% accurate, respectively.

For the two AlexNet instances, namely AlexNet-TH(ick) and AlexNet-SW(ell), 
1) we apply the greedy channel selection and 
train {\FGAN} with the twenty selected channels,
2) generate test images with the channel intensity from 0.33 to 3 times of the original, and 3) measure the unexpectedness scores for the channels;
we rank by the score the test data for the channels and examined top-5 channels' test data in detail.
For both AlexNet-TH and AlexNet-SW, we noticed that for one particular channel,
namely thick channel and swollen channel,
its test data result in incorrect inference outcome at a high rate.
Table\,\ref{tbl:defect-stat} shows the rate of the generated test data for thick and swollen channels
resulting in mis-classification for varying intensity input $I$; the table also shows the  
ranking of those channels by the score in parentheses in the header.

\begin{figure}[t]
    \includegraphics[width=\linewidth]{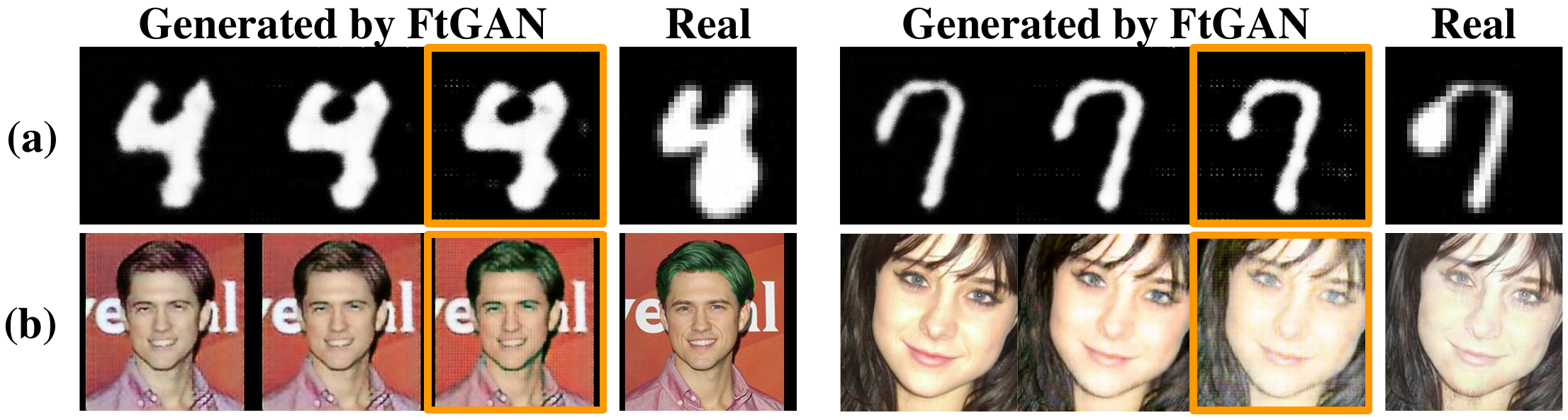}
    \caption{
    Testing defective CNN instances that misclassify (a) swollen digits
    or (b) faces with green hair or pale skin.
    }\label{fig:debug-result}\par
\end{figure}

Fig.\,\ref{fig:debug-result}\,(a) shows the images that are generated for AlexNet -SW for the swollen channel.
The generated images with three different intensity are shown for each digit; images in Morpho-MNIST are also shown (marked as ``Real'').
The generated images have swollen strokes like those in Morpho-MNIST.
The swelling is subtle in our test data but it is sufficient 
for the classifier to output incorrect labels;
the orange boxes in Fig.\,\ref{fig:debug-result} indicate 
that the samples are incorrectly classified by AlexNet-SW.
The supplementary material has more comprehensive analysis including the process of
finding and analyzing the defects.

\begin{figure}[tb]
    \includegraphics[width=\linewidth]{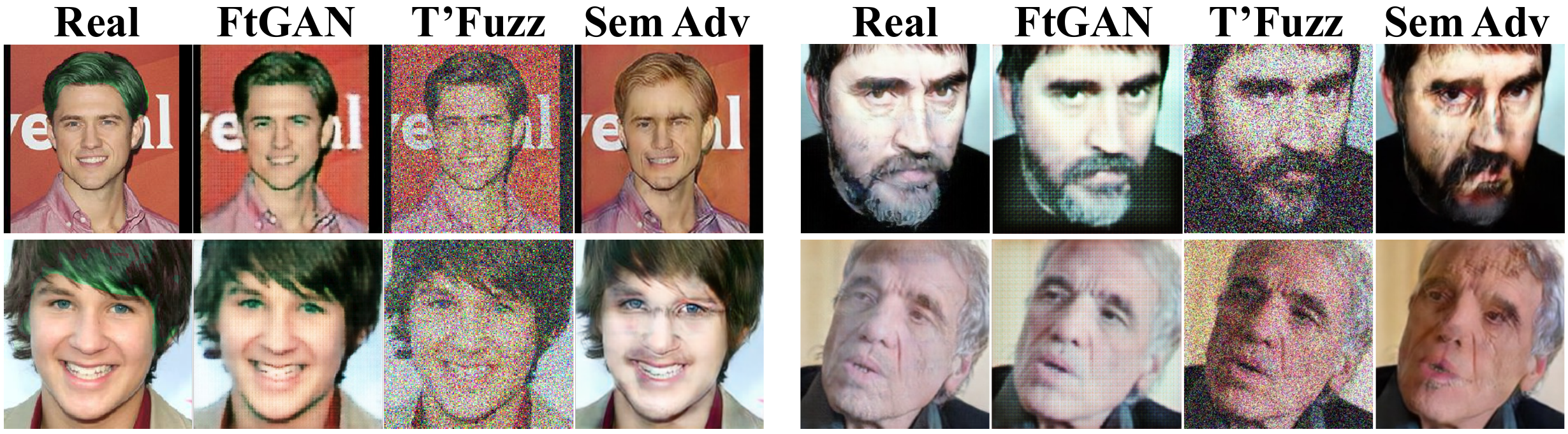}
    \caption{
    Test images by {\FGAN}, TensorFuzz, and semantic adv attack
    for VGG-hair\,(left) \& ResNet-skin\,(right).
    }\label{fig:debug-cmp}
\end{figure}

\begin{figure}[t]
    \centering
    \includegraphics[width=0.58\linewidth]{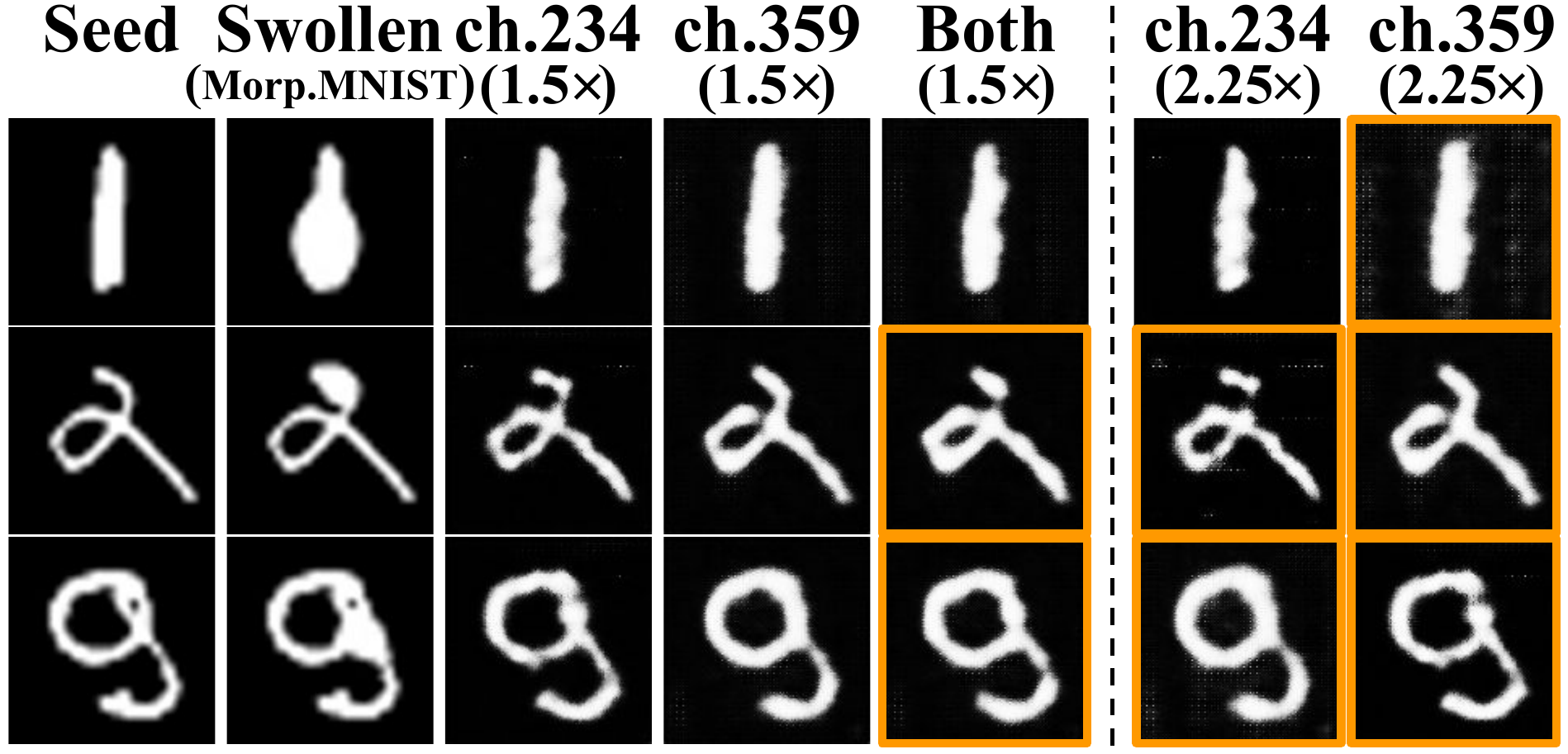}
    \captionof{figure}{Testing of two channels in AlexNet-SW.}
    \label{fig:multi-ch-test}\par
\end{figure}

\begin{table}[b]
    \scriptsize
    \centering
    \setlength\extrarowheight{-1.25pt}
    \caption{
    The rate of mis-classified test data generated for the defective channels.
    The rankings of the channels by their unexpectedness scores are shown in parentheses in the header.
    }
    \begin{tabular}{|c|ccc|}
    \hline
    \multirow{2}{*}{\textbf{\begin{tabular}[c]{@{}c@{}}Range of \\ Intensity $I$\end{tabular}}} & \multicolumn{3}{c|}{\textbf{Model (score rank / tested channels)}} \\ \cline{2-4} 
     & \multicolumn{1}{c|}{\textbf{\begin{tabular}[c]{@{}c@{}}AlexNet-TH (2/20)\end{tabular}}} & \multicolumn{1}{c|}{\textbf{\begin{tabular}[c]{@{}c@{}}AlexNet-SW (4/20)\end{tabular}}} & \textbf{\begin{tabular}[c]{@{}c@{}}VGG-hair (3/40)\end{tabular}} \\ \hline
    0.9 - 2.0 & \multicolumn{1}{c|}{31.8\%} & \multicolumn{1}{c|}{16.8\%} & 56.7\% \\ \hline
    0.7 - 3.0 & \multicolumn{1}{c|}{67.8\%} & \multicolumn{1}{c|}{56.2\%} & 80.4\% \\ \hline
    0.5 - 4.0 & \multicolumn{1}{c|}{84.8\%} & \multicolumn{1}{c|}{85.2\%} & 87.4\% \\ \hline
    0.4 - 4.5 & \multicolumn{1}{c|}{89.0\%} & \multicolumn{1}{c|}{90.2\%} & 89.4\% \\ \hline
    \end{tabular}
    \label{tbl:defect-stat}
\end{table}

We also tested the interactions of two channels of AlexNet-SW by feeding the output of
{\FGAN} that is trained for one channel as the input of another {\FGAN} that is trained for a different channel.
We tested the pairs of top-5 channels by their unexpectedness scores.
Fig.\,\ref{fig:multi-ch-test} shows the results with comparing the test data generated for 
both channels (denoted by {\it Both})
with the data generated for one channel (denoted by {\it ch.} followed by the channel id).
The data is generated with 1.5$\times$ channel intensity; data generated with 2.25$\times$ channel intensity is also shown for comparison.
The orange box denotes that the sample is mis-classified by AlexNet-SW.
From the results, we observed that for some seed inputs the swelling is more noticeable
(hence more mis-classifications) when the intensities of both channels are adjusted.
Although our focus in this paper is modular testing of individual CNN channels, 
this experiment shows that {\FGAN} can be used to jointly test multiple CNN channels.
The supplementary material has more results and discussions.

Moreover, we simulated the data corruption bugs with the VGG Face dataset.
We trained two VGG-16 networks to identify the faces in the dataset but we injected 
certain bugs.
For one instance (namely VGG-hair) we trained it 
to misclassify any faces with green hair as a certain target person;
for another one (namely VGG-skin) we made it to
infer any faces with pale skin as another target identity.
We have repeated the process in a similar manner
to train two faulty ResNet instances (ResNet-\{hair,skin\}).
For ResNet though, we only considered 3x3 convolutional layers for the testing
as 1x1 convolutions are mainly designed to control the computational complexity 
by reducing or expanding the channel dimensions\,\cite{szegedy2015going,he2016deep}.

To test the faulty CNN instances, we selected forty channels (or sixty for ResNet) with the greedy selection in Algorithm~\ref{alg:selection} and tested them with {\FGAN}. We examined the test data of top-5 unexpected channels
in detail and identified one most defective channel for each of 
the instances. 
Let us describe the details of the manual examination for the VGG models.
We describe the case for VGG-skin, and the case for VGG-hair is described in the supplementary material.
Among the five channels with high unexpectedness scores,
four channels are related to human-recognizable attributes (nose-color, liver-spots, pale-skin-a, and pale-skin-b); we refer these channels as {\it semantic} channels.
The fifth channel adds grid patterns to the images.
For the five channels, we examined how the inference outcome changes as the intensity changes. 
The inference outcomes for all five channels changed in a biased manner
towards the target identity that VGG-skin is trained for.
For the three semantic channels (except the pale-skin-b channel), 
20--30\% of the incorrect outcomes are inferred as the target identity.
For the pale-skin-b channel, 60\% of the incorrect outcomes are inferred as 
the target identity.
Fig.\,\ref{fig:debug-result} shows an example of generated test data.
Again with our testing techniques we successfully identified the defect in VGG-skin.
For other channels having low unexpectedness scores among the forty selected ones, we did not observe such biased inference changes.

If we test the faulty models with TensorFuzz or semantic adversarial attack\,\cite{joshi2019semantic}, the defects are not detected.
Fig.\,\ref{fig:debug-cmp} shows the generated images of {\FGAN} and the two techniques.
TensorFuzz generates noise-augmented images; the semantic attack inconsistently changes a few attributes.
Most importantly, the images made by the two techniques are not identified as the target person,
thus they did not find the defect but simply generated adversarial examples.
We also applied other adversarial example generation techniques\,\cite{carlini2017towards,moosavi2016deepfool},
but they generated minimally perturbed images (that are not identified as the target person) and thus we do not show them here.
Furthermore, we used Grad-CAM\,\cite{selvaraju2017grad},
the state of the art XAI technique, to \FGAN's test images
and confirmed that the changes made by {\FGAN} caused the inference changes (discussed in the supplementary material).

\subsection{Finding Bugs in a Public CNN Model}
\label{sec:eval-public-cnn}

\begin{figure}[bt]
    \centering
    \includegraphics[width=0.9\linewidth]{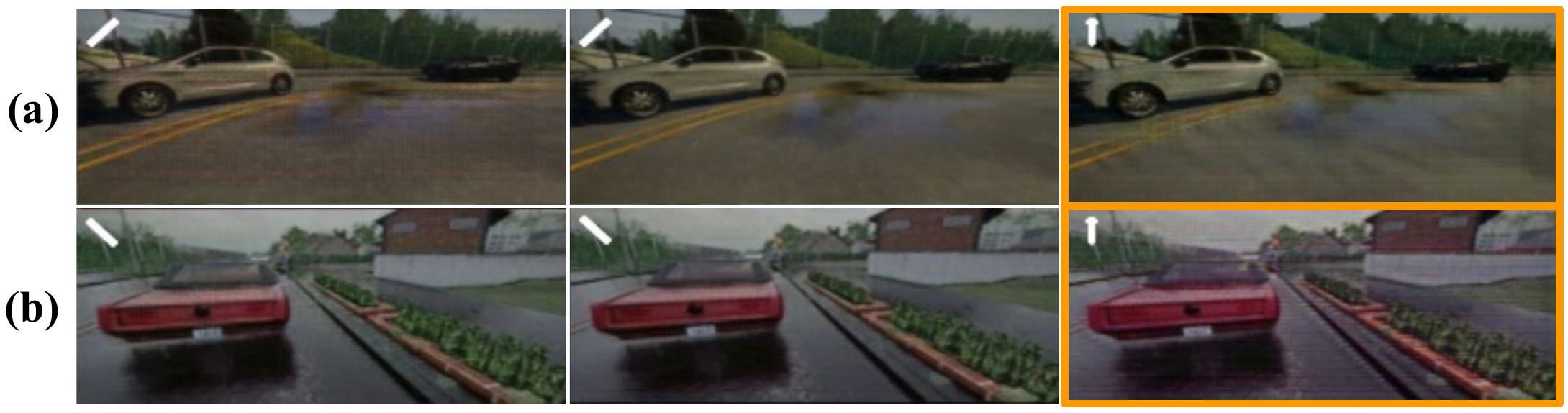}
    \caption{
    Test images generated by {\FGAN} for CARLA-CNN. 
    We observe (a) the center line wear-out and road texture changes, and (b) the color tone changes.
    }\label{fig:debug-carla}\par
\end{figure}

We further evaluated {\FGAN} with a pre-trained, publicly-available CNN instance 
for autonomous driving that is developed and trained by Codevilla et al.\,\cite{carla}.
This CNN instance, which we call CARLA-CNN, has eight convolution layers and two dense layers. 
We tested the channels of the convolution layers in the same way as the previous experiments.
The images generated for a subset of the channels have certain semantic variations 
such as the center line wear-out, road texture changes, or color tone changes.
Fig.\,\ref{fig:debug-carla} shows the generated images; these channels generally have high unexpectedness scores.
The white arrows on the images are the steering decision made by CARLA-CNN.
We can see that the variations caused by the intensity changes
make CARLA-CNN to make wrong steering decision (marked by orange boxes). The supplementary material
has more discussion on the experimental results.

\begin{figure}[bt]
    \centering
    \includegraphics[width=0.9\linewidth]{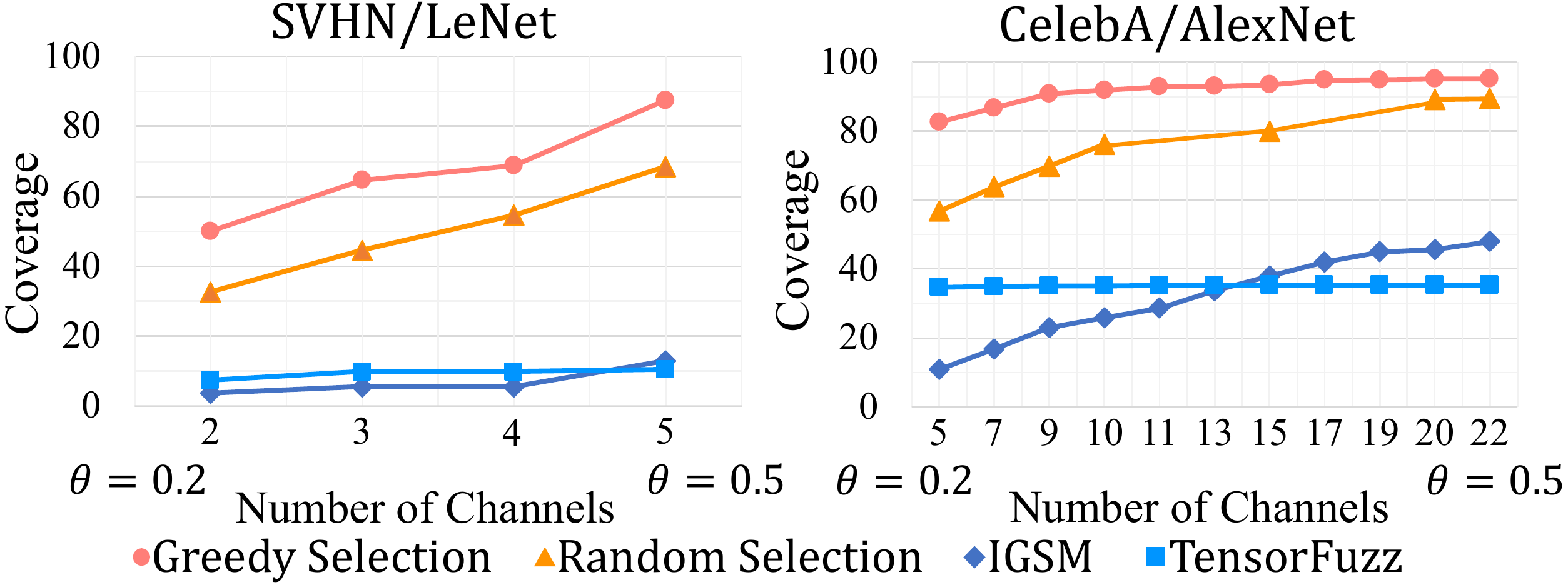}
    \caption{Channel boundary coverage of {\FGAN} (greedy \& random selection), IGSM, and TensorFuzz. $\theta$ below the x-axis is the correlation threshold for the channel selection. 
    }\label{exp:coverage}\par
\end{figure}

\subsection{Coverage Gain by Greedy Channel Selection}
\label{sec:eval-coverage}

\noindent {\bf Metric for channel coverage.}
With a test coverage metric, we aim to quantify the fraction of the channels in a target CNN
that satisfy the test condition for a test suite made by a test image generator.
Borrowing the concept of neuron boundary coverage \cite{ma2018deepgauge}, we define {\em channel boundary coverage}. 
Let $T_{c}(x)$ be the intensity of channel $c$ when image $x$ is given as input to target network $T$. 
Given training dataset $\mathbb{X}$ for $T$, let $I_{c}^{upper}$ be the maximum intensity of the channel $c$ for $\mathbb{X}$; i.e., $I_c^{upper}$$=$$\max_{x \in \mathbb{X}} T_c(x)$.
For the channel $c$, if $T_c(x)$ for a test image $x$ from the test suite $\mathbb{T}$ 
is larger than $I_c^{upper}$, the boundary of $c$ is said to be {\em covered} by $\mathbb{T}$. 
Let $V$ denote the set of all channels to be tested in $T$. 
The {\em boundary coverage} is then defined as $\frac{|\{ c \in V | \exists x\in \mathbb{T}, T_c(x) > I_c^{upper} \}|}{|V|}$, that is, the ratio of channels whose boundaries are covered by at least an image in $\mathbb{T}$.

\vspace{2mm}
\noindent {\bf Coverage evaluation.} With varying the minimum correlation $\theta$ from 0.2 to 0.5 in Algorithm~\ref{alg:selection}, we selected test channels of the target networks trained for SVHN (LeNet) and CelebA (AlexNet).
For the baselines, we selected the same number of test channels with random selection. 
For the two networks, we then generated $2,000$ test images per channel based on the seed dataset using {\FGAN} and plotted the boundary coverage of each test in Fig.\,\ref{exp:coverage}.
We also applied two other techniques, IGSM (adversarial example generation based on iterative gradient sign method)\,\cite{kurakin2016adversarial} and TensorFuzz, to generate the same number of test images for each experiment; we cannot apply semantic adversarial attack as its execution time is too long,
taking more than a few seconds to generate a single image.
The graphs show that the test suite obtained by our greedy algorithm achieves the highest coverage in both networks
and verify that the channels selected by our greedy algorithm efficiently cover for the untested channels.


\section{Conclusion}
\label{sec:conclusion}

This paper proposes techniques for testing the channels of CNNs.
We designed {\FGAN}, an extension to GAN that generates realistic test data for a target CNN
with varying its selected channels intensities.
We developed a channel selection algorithm that finds a subset of a CNN's representative channels
by using the correlations between the channels.
To investigate inconsistency in the target CNN's inference with \FGAN's test data,
we defined unexpectedness score and rank the test data by this score.
In our evaluation, we investigated five CNN models that are trained with five datasets. 
By applying our testing techniques, we successfully found defects in both synthetic and real-world CNN instances.

\section{Acknowledgement}
\label{sec:Acknowledgement}

This work was supported by the National Research Foundation of Korea (NRF) grant funded by the Korea government (MSIT) (No.2018R1D1A1A02086132 and No.2020R1G1A1011471) and by
Institute of Information \& communications Technology Planning \& Evaluation (IITP) grant funded by the Korea government (MSIT) (No.2020-0-01373, Artificial Intelligence Graduate School Program (Hanyang University)).
We thank Nahun Kim for his help with the experiments.
Jiwon Seo is the corresponding author.

\bibliography{ftgan}



\clearpage
\appendix

\counterwithin{figure}{section}
\counterwithin{table}{section}

\setcounter{section}{0}
\setcounter{figure}{0}

\setcounter{table}{0}

\section{Supplementary Material for Paper (Testing the Channels of Convolutional Neural Networks)}

This section supplements the evaluation section of the paper (Section\,\ref{sec:eval}).
Specifically, Section\,\ref{sec:exteval1} extends Section\,\ref{sec:eval-feature},
\ref{sec:exteval2} extends \ref{sec:eval-find-problem}, and
\ref{sec:exteval3} extends \ref{sec:eval-public-cnn}.

\subsection{Realistic Test Generation and Unexpectedness Score (Extension of Section\,\ref{sec:eval-feature})}
\label{sec:exteval1}

\noindent
{\bf Experiments with MNIST, SVHN, and CIFAR-100.}
We first discuss the validity of unexpectedness score with the MNIST and CIFAR-100 datasets.
Specifically we look into the channel correlations and discuss if the values may represent
the corresponding inference computations.
For the investigation, we trained three AlexNet instances with the MNIST, SVHN, 
and CIFAR-100 datasets.
We first examined the correlations of the channel intensities in each layer
(a) using the entire datasets and (b) using only the data in each class.
We noticed that for quite a few channel pairs, the correlations computed for (a) and (b) 
are substantially different. 
When we counted the number of channel pairs in the forth layer for which 
the correlation difference (computed for (a) and (b)) is considerably large (larger than 0.2),
the counts are approximately one hundred for the SVHN and CIFAR-100 models and 
one thousand for the MNIST model; the numbers for the channels in other layers is similar.
The maximum correlation differences are  0.55 (from 0.4 to 0.95) for MNIST,
0.35 (from 0.41 to 0.76) for SVHN, and 0.3 (from 0.41 to 0.71) for CIFAR-100.

\begin{figure*}[htb]
\centering
\parbox{0.65\columnwidth}{
    \centering
    \includegraphics[width=\linewidth]{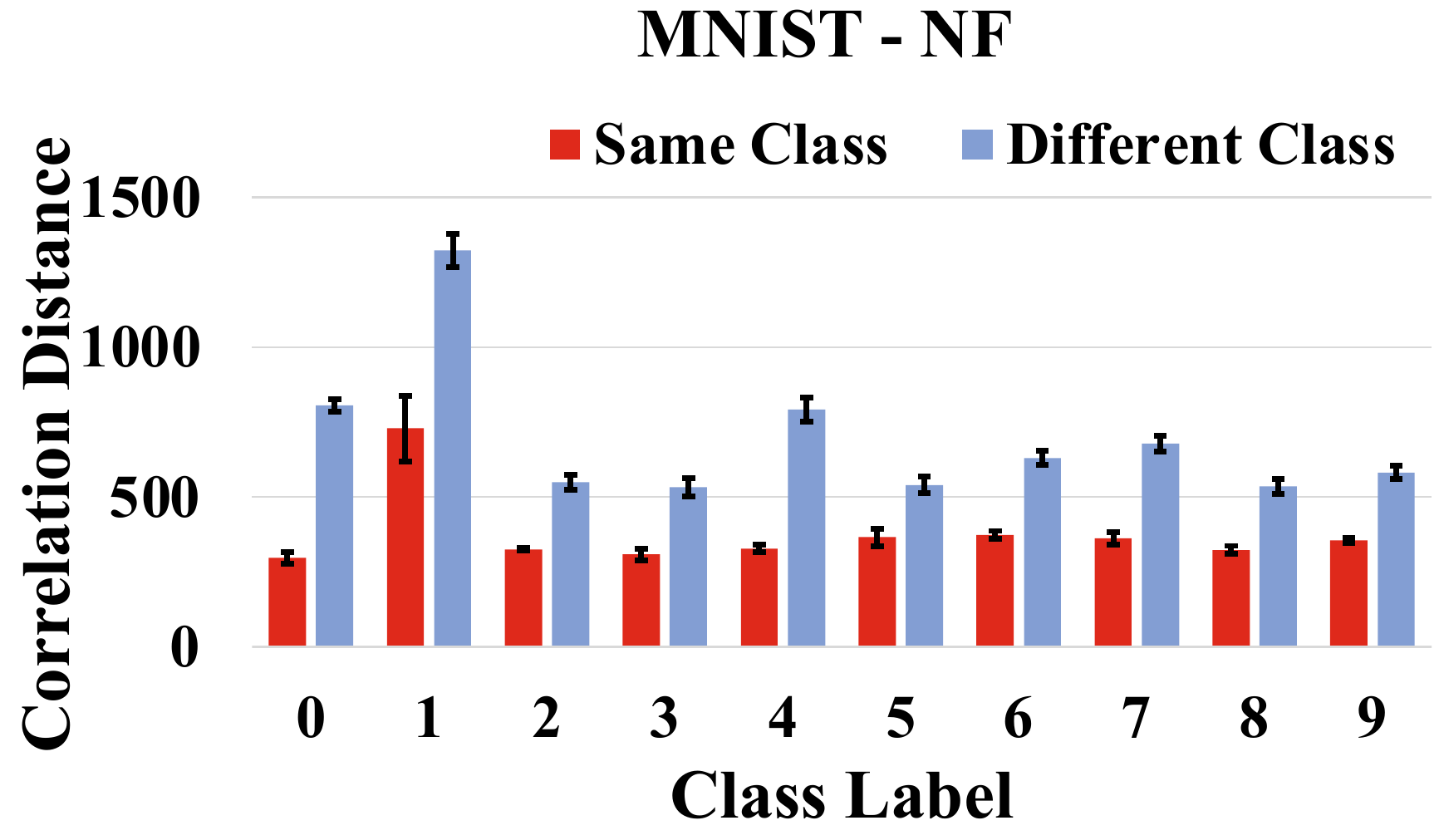}
    \caption{L1 distances between the sub-groups of same and different classes in MNIST-NF.}
    \label{fig:l1_mnist_orifrac}
}
\hspace{2mm}
\parbox{0.65\columnwidth}{
    \centering
    \includegraphics[width=\linewidth]{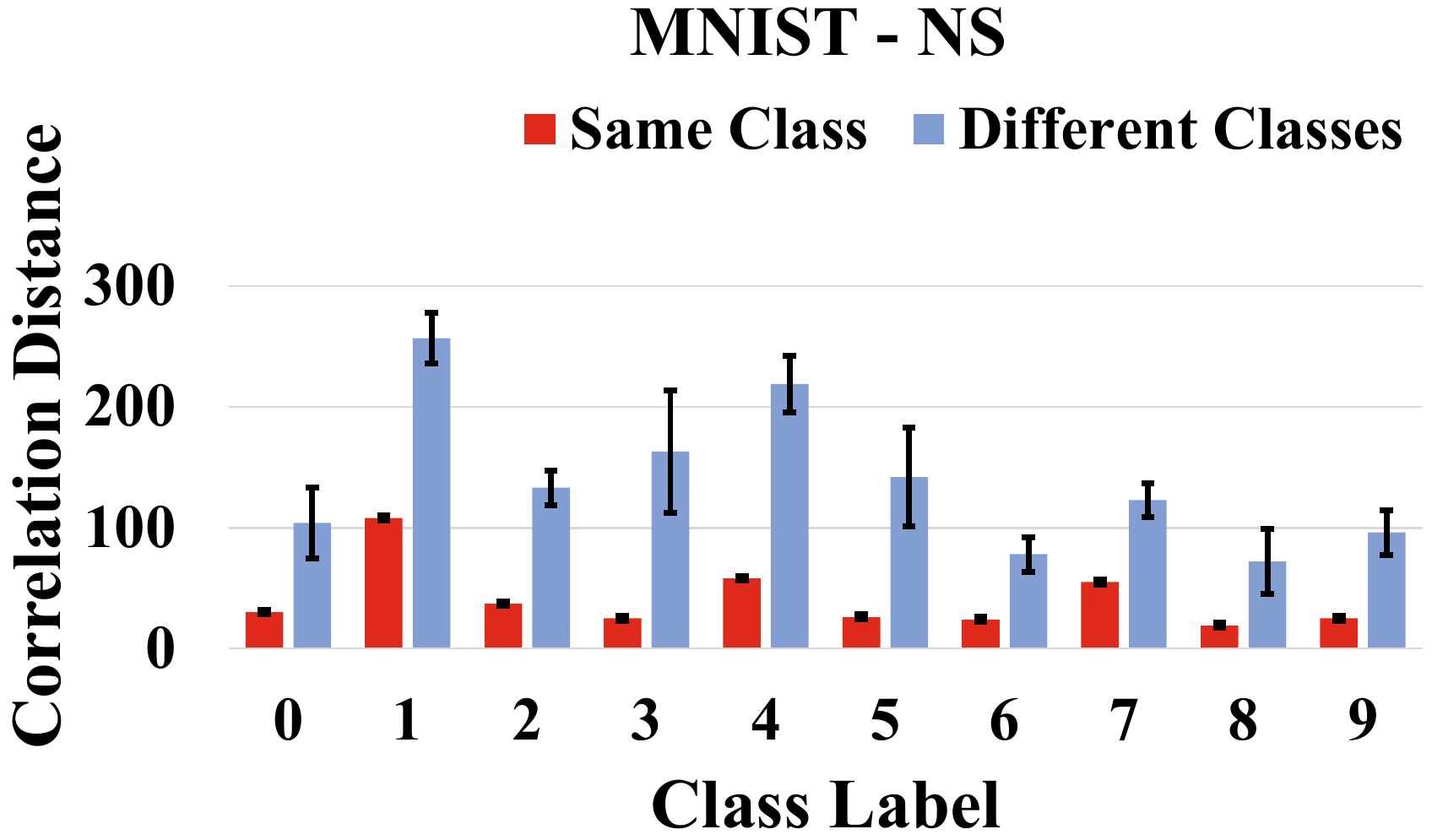}
    \caption{L1 distances between the sub-groups of same and different classes in MNIST-NS.}
    \label{fig:l1_mnist_oriswell}
}
\hspace{2mm}
\parbox{0.65\columnwidth}{
    \centering
    \includegraphics[width=\linewidth]{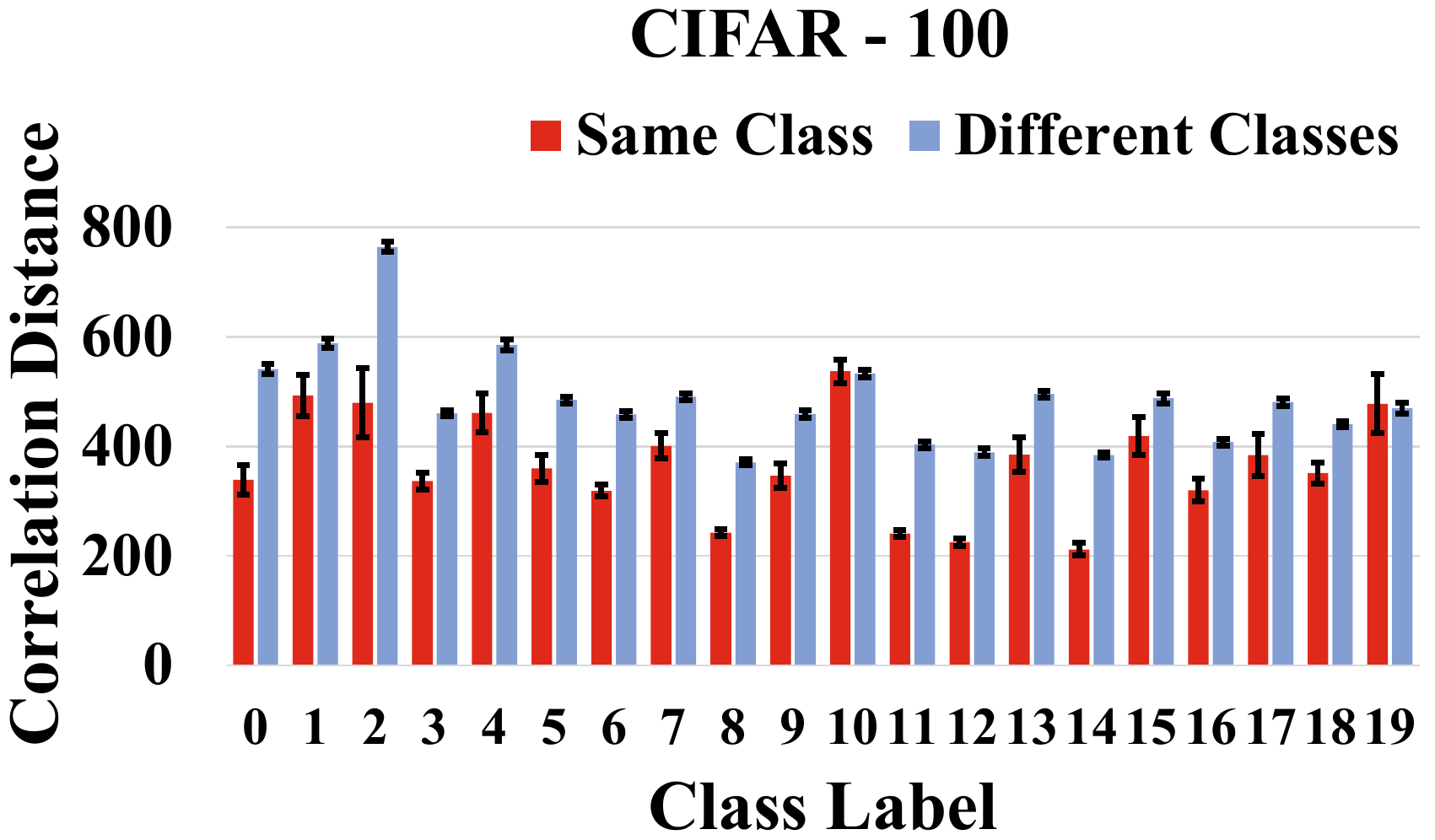}
    \caption{L1 distances between the sub-groups of same and different classes in CIFAR-100.}
    \label{fig:l1_cifar100}
}
\end{figure*}

The experimental results show that many channel pairs have class-selective
strong correlations, from which we conjecture that the channel correlations
for the data of a same class may represent CNN's inference computations for those data.
That is, the channel correlations for the data of a same class may be considered as
a {\it characteristic} of the inference computations for those data.
This is fundamentally equivalent to the previous results of applying CCA for 
the analysis of hidden tensors\,\cite{svcca2017}, but we examined at a feature level.
To further investigate the channel correlations and inference computations, 
we sub-grouped {\it similar} data within a same class and examined their channel correlations. Specifically, we created two new datasets, namely, MNIST-NF and MNIST-NS from Morpho-MNIST
such that MNIST-NF consists of normal and fracture digits and MNIST-NS consists of normal and swollen digits.
We considered those normal and fracture digits (or normal and swollen digits) in each class as a sub-group;
e.g. normal 2's and fracture 2's are two sub-groups of all the 2's.
Then we measured the L1 distances between each sub-group's top-5\% channel correlations.
The measured distances are plotted in Fig.\,\ref{fig:l1_mnist_orifrac} (for MNIST-NF) and Fig.\,\ref{fig:l1_mnist_oriswell} (for MNIST-NS).
The red bars are the correlation distances between the sub-groups within a same class and the blue bars
are the distances between the sub-groups in different classes; the error bars
are the standard errors.
The figures show that the distances between similar sub-groups in a same class
are much closer than the distances between dissimilar sub-groups in different classes.

We repeated a similar experiment with the CIFAR-100 dataset that has 20 super-classes and
100 sub-classes (i.e., five sub-classes in each super-class). 
We trained an AlexNet instance to classify the 20 super-classes
and computed the channel correlations for each of the 100 sub-classes.
Then we measured the L1 distances between the correlations of the sub-classes and plotted the results in Fig.\,\ref{fig:l1_cifar100}.
The sub-groups generally have smaller L1 distances when they are in a same super-class compared to the cases
when they are in different super-classes. For the sub-groups in 18 super-classes (out of 20), 
their distances to the sub-groups in a same super-class are shorter than the distances to the sub-groups in  different super-classes.
For the sub-groups in the other two super-classes (super-class 10 and 19), 
their distances to the sub-groups in other super-classes are slightly shorter.
When we examined those super-classes, their sub-classes seem to have a larger diversity
compared to other super-classes. For example, super-class 10 is for ``large natural outdoor scenes''
and its sub-classes are  ``cloud'', ``forest'', ``mountain'', ``plain'', ``sea'';
the sub-classes have very diverse shapes, colors, and textures.
We examined all the other super-classes and their sub-class diversities; 
if the sub-classes are similar (in shapes, colors, and textures), their L1 distance between each other
is generally shorter.

From the experiments we concluded that the (top-k) channel correlations for a similar group of data (such as 
the data in a same class) represent, to some extent, the inference computations for that group of data.
Thus we propose to use the top-k channel correlations as a reference point of CNN's inference computations.
Moreover, we measure the L1 distance between the correlations computed for training data
and that computed for test data and use the distance as a metric of {\it unexpectedness} of the 
inference computations for the test data.
Note that we do not claim that the unexpectedness metric is highly accurate.
When we apply the unexpectedness metric to test and debug CNN's features,
we also use other information, such as the generated test data or the inference outcome with the test data,
to understand the test results; thus the metric does not need to be absolutely accurate.

\begin{figure}[h]
    \centering
    \includegraphics[width=\linewidth]{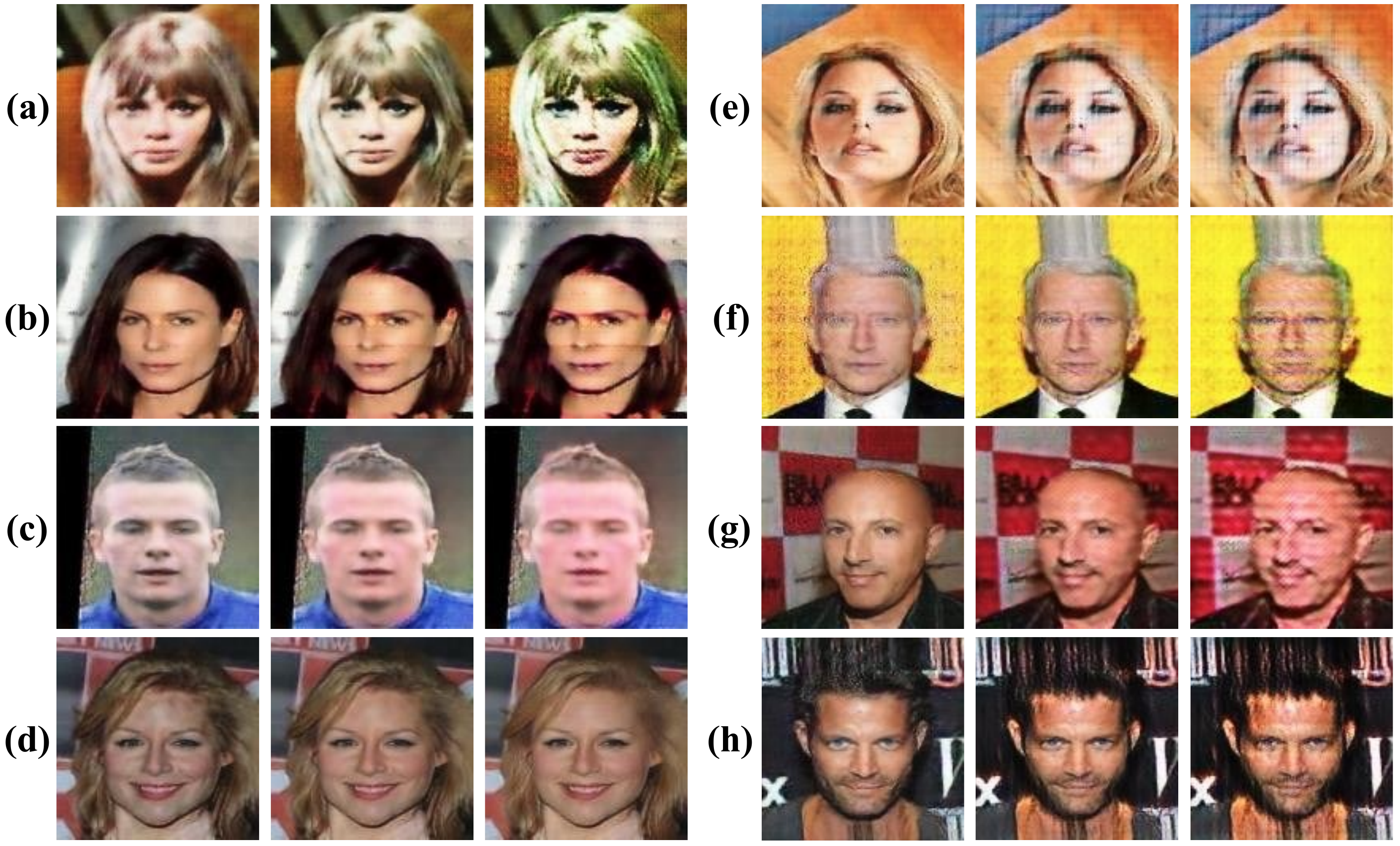}
    \caption{(Extension of Fig.\,\ref{fig:realistic-data-gen}) 
    The images are generated for the CelebA model.
    The human-recognizable variations are (a)hair color, (b)face mask, (c)face color and (d) age;(e)--(h) are not human-recognizable.}
    \label{fig:ext_fig3_celeba}
\end{figure}

\noindent
{\bf Experiments with CelebA and VGG Face.}
Now we discuss the realistic test generation and validity of unexpectedness score
with the CelebA and VGG Face datasets; 
this is described in Section\,\ref{sec:eval-feature} but we provide
more comprehensive results and discussion here.
Fig.\,\ref{fig:ext_fig3_celeba} shows \FGAN's generated images grouped by the tested channels for the CelebA model; the figure extends Fig.\,\ref{fig:realistic-data-gen} in the manuscript. 
In addition to the pale-skin attribute interfering the age classifier, we
noticed a few more interesting cases.
For example, as we change the intensity of the hair color channel, some of the {\it pale-skin} data 
are incorrectly inferred as not {\it pale skin}, even though the skin color 
changes little to none; an example of this is in Fig.\,\ref{fig:realistic-data-gen} in the paper.
Also, the face mask channel seemed to confuse quite a few other classifiers for the data in  {\it pale-skin}, {\it not pale-skin}, {\it young}, {\it not brown hair}, and other classes; example images are in Fig.\,\ref{fig:realistic-data-gen} (b) in the paper and also in Fig.\,\ref{fig:ext_fig3_celeba} (b) here.
The images in (a)--(d) of Fig.\,\ref{fig:ext_fig3_celeba} are generated to test the same channels in Fig.\,\ref{fig:realistic-data-gen}.
That is, the images in (a) are generated for the same channel for the images in Fig.\,\ref{fig:realistic-data-gen} and the same for (b)--(d).
We can see that varying the intensity of a same channel transforms 
the corresponding four images in a similar way. The images in (e)--(h) (with non human-recognizable attributes)
are generated for the same channels in (e)--(h) of Fig.\,\ref{fig:realistic-data-gen}.

\begin{figure}[th]
    \centering
    \includegraphics[width=\linewidth]{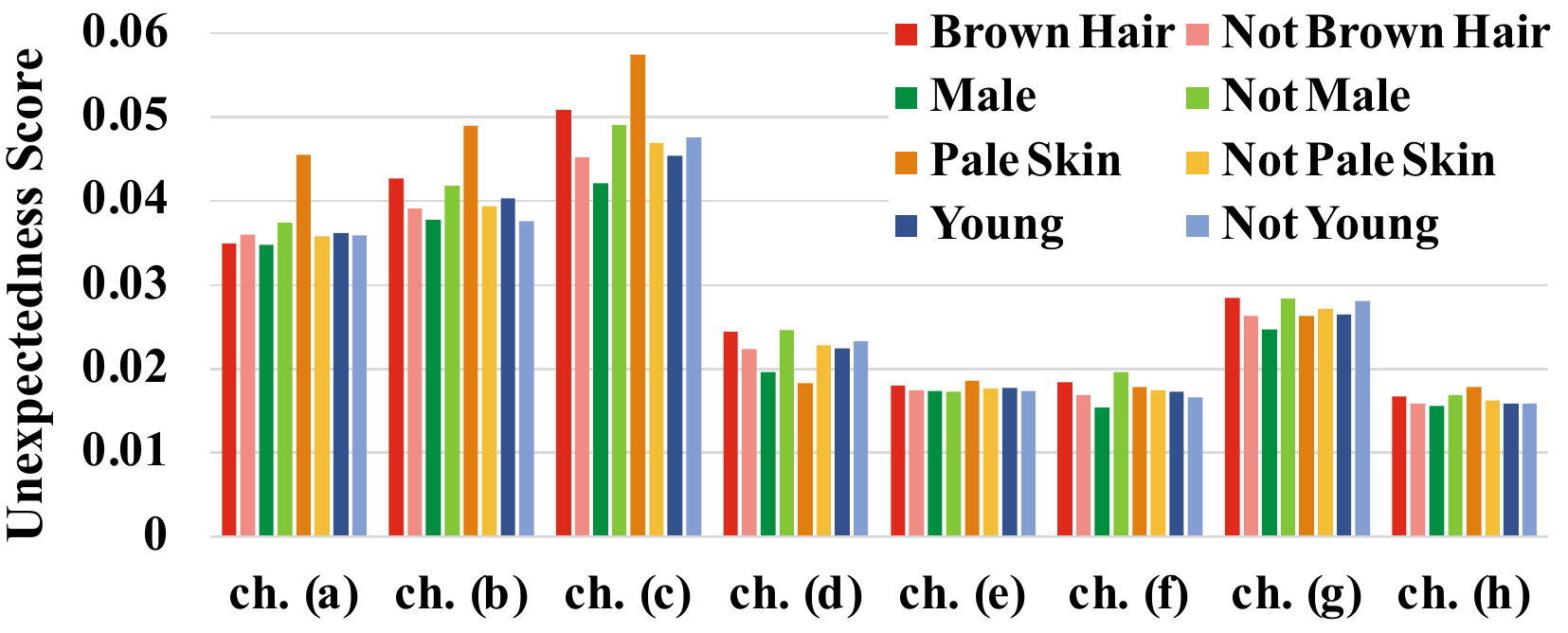}
    \caption{Normalized unexpectedness scores for the channel (a)--(h) in Fig.\,\ref{fig:realistic-data-gen} (or Fig.\,\ref{fig:ext_fig3_celeba}).}
    \label{fig:unexp-scores-extension-celeba}
\end{figure}

\noindent
{\bf Validity of Unexpectedness Score.} For the generated test data in Fig\,\ref{fig:realistic-data-gen},
we also measure and examine their unexpectedness scores.
Fig.\,\ref{fig:unexp-scores-extension-celeba} plots the unexpectedness scores of the channels for the attributes a--h in Fig.\,\ref{fig:realistic-data-gen} for a subset of the classes in the dataset.
The channels for the human-recognizable attributes show higher unexpectedness scores
(which is discussed next)
except for the age channel, or ch.(d) in Fig.\,\ref{fig:unexp-scores-extension-celeba} and Fig.\,\ref{fig:realistic-data-gen}(d).
When we examined the test data for the age channel, the changes of the attribute and their inference computation
seemed reasonable.
That is, as we change the intensity of the age channel, the inference outcome for some data in {\it not-male} classes changed to {\it male};
however, the inference changes are limited to the face images that seemed gender neutral (such as Fig.\,\ref{fig:realistic-data-gen}(d) right),
for which deciding their gender is harder with age transformations.
For the other three channels (hair color, face mask, and face color channels),
we observed several inconsistent behaviors.
For example, testing the face color channel with the {\it pale-skin} class data
generated pink-ish face data, which confused the classification of other attributes; 
i.e., the generated test data is classified as not {\it young}
even though the data is labeled as {\it young} and the seed image is classified as such.
Fig.\,\ref{fig:realistic-data-gen}(c) (right) shows an example of such cases; with the pink-ish skin she is 
(incorrectly) classified as not {\it young}. 
Also, the hair color channel shows similar behavior and confused {\it pale-skin} classifier.
In contrast, we observed that non-recognizable channels (e--h in Fig.\,\ref{fig:realistic-data-gen})
have relatively low unexpectedness scores. 
When we examined their inference computation with different intensity input,
the intensity of the tested channels (and some related channels) varied accordingly,
but the pair-wise channel correlations are not affected much, hence the low unexpectedness scores.
We mainly examined the human-recognizable channels as their analysis is more straightforward;
it requires further study to understand how the non-recognizable channels affect the inference computation.
Here we showed that unexpectedness score helps to find inconsistent inference computations for generated test data.

\begin{figure}[th!]
    \centering
    \includegraphics[width=\columnwidth]{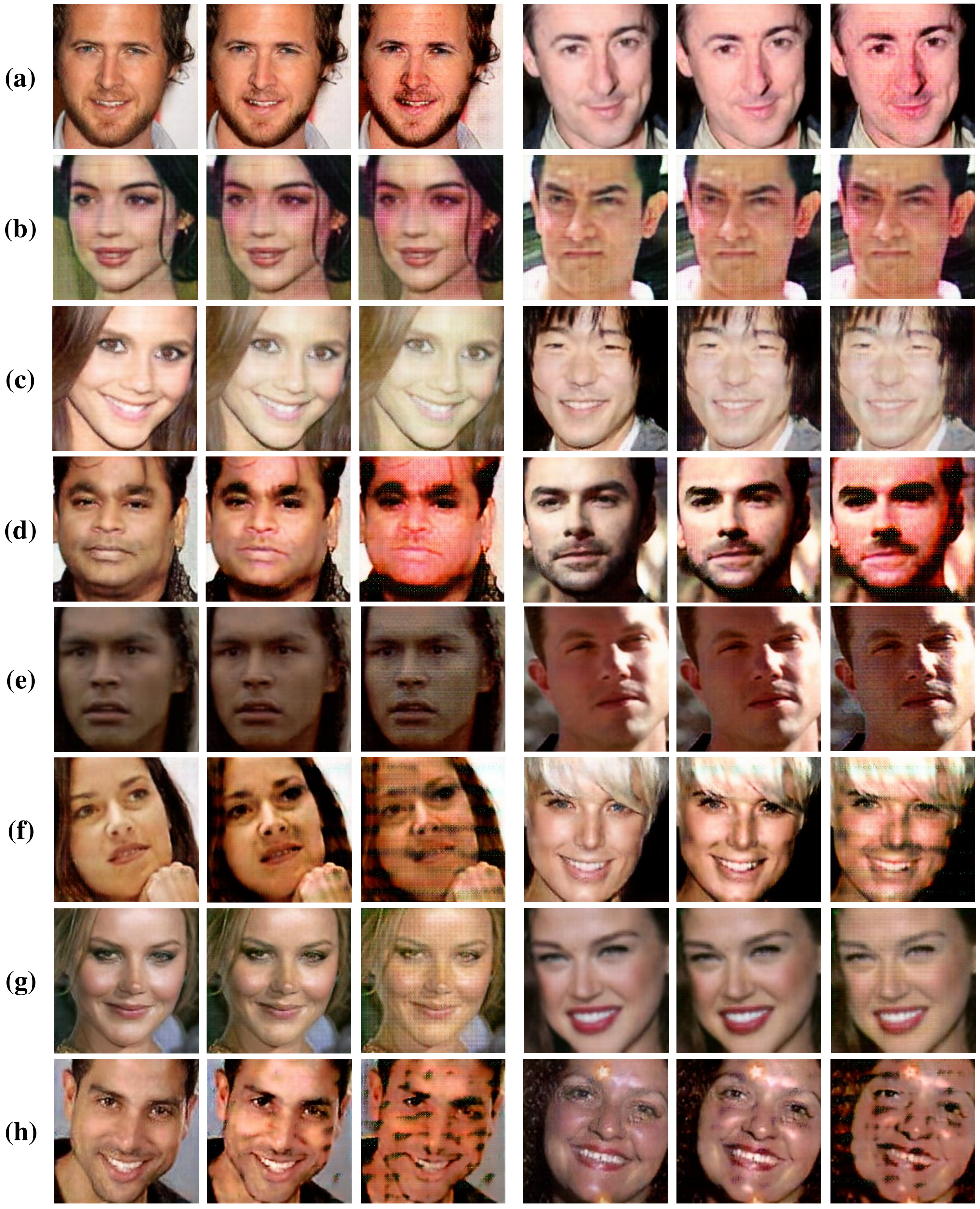}
    \caption{(Extension of Fig.\,\ref{fig:realistic-data-gen}) 
    The images are generated for the VGG Face model. The human-recognizable variations are (a)\,red nose, (b)\,face makeup, (c)\,pale skin and (d)\,face color; the images in (e)--(h) 
    have non human-recognizable variations.}
    \label{fig:ext_fig3_vgg}
\end{figure}

Fig.\,\ref{fig:ext_fig3_vgg} also extends Fig.\,\ref{fig:realistic-data-gen} but the images are generated for the VGG Face model.
Similarly as in the CelebA model, we observed human-recognizable variations (a--d) and
non human-recognizable variations (e--h). 

\begin{figure}[t!]
    \centering
    \includegraphics[width=\linewidth]{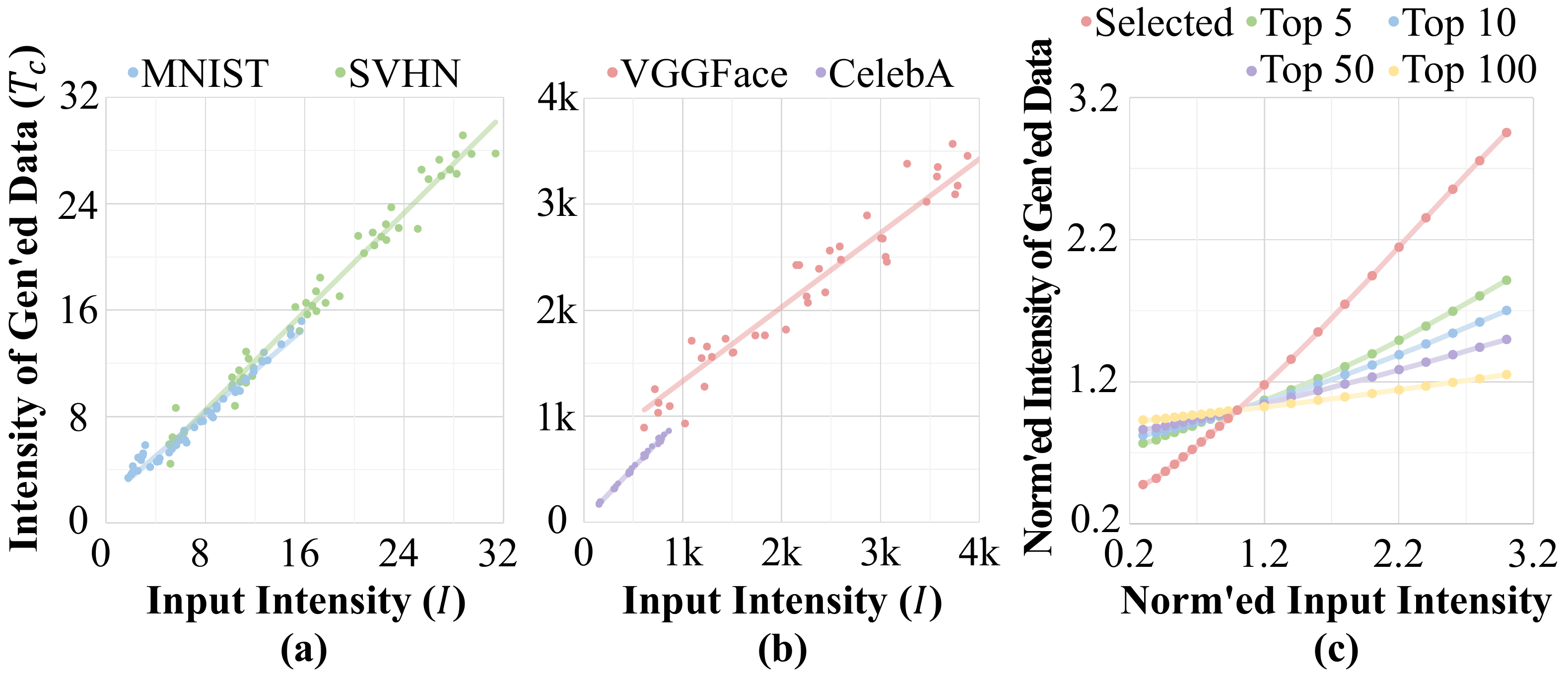}
    \caption{\FGAN's intensity input $I$ and the actual channel intensities of the generated images;
    (a) is MNIST/SVHN and (b) is CelebA/VGG\,Face.
    The plot (c) is the intensities of a tested channel and other channels for a sample image. }
    \label{fig:intensity_plots}
\end{figure}

To evaluate whether {\FGAN} can control the channel intensity ($T_c$) of the generated images,
we trained {\FGAN} to test randomly-selected channels, or features, of the four CNN models with their corresponding training datasets.
Then we generated test data with {\FGAN} using the images in the test sets
as the seeds. We varied the intensity input $I$ to be between 0.33 and 3 times that of the seeds and observe the channel intensities of the generated images.
The results are plotted in Fig.\,\ref{fig:intensity_plots}\,(a) and (b) grouped by the digit and face datasets.
As the plot shows, the images generated by {\FGAN} vary the intensity of the tested channel in the target CNN 
according to its input $I$.
Fig.\,\ref{fig:intensity_plots}(c) shows the normalized intensity changes 
of the selected channel and other channels for a sample seed image.
We can see that the selected channel (and a few correlated ones denoted by Top\,5 and Top\,10)
are controlled by input $I$, but the intensities of other channels
are not affected much.

\subsection{Finding Data Corruption Bugs (Extension of Section\,\ref{sec:eval-find-problem})}
\label{sec:exteval2}

\noindent
{\bf Analysis of AlexNet-SW.}
We first present a comprehensive analysis of the AlexNet-SW experiment in Section\,\ref{sec:eval-find-problem}.
We trained {\FGAN} for AlexNet-SW with twenty selected channels and measured their unexpectedness scores.
The results are shown in Table\,\ref{tbl:swell-unexpectedness}. 
Because the number of top-5\% channel correlation pairs is different 
from one layer to another, we present the scores in a normalized form as well.
Then we examined the generated test data for the top-5 channels by the raw and normalized unexpectedness scores.

\begin{figure}[th!]
    \centering
    \includegraphics[width=\linewidth, height=105mm]{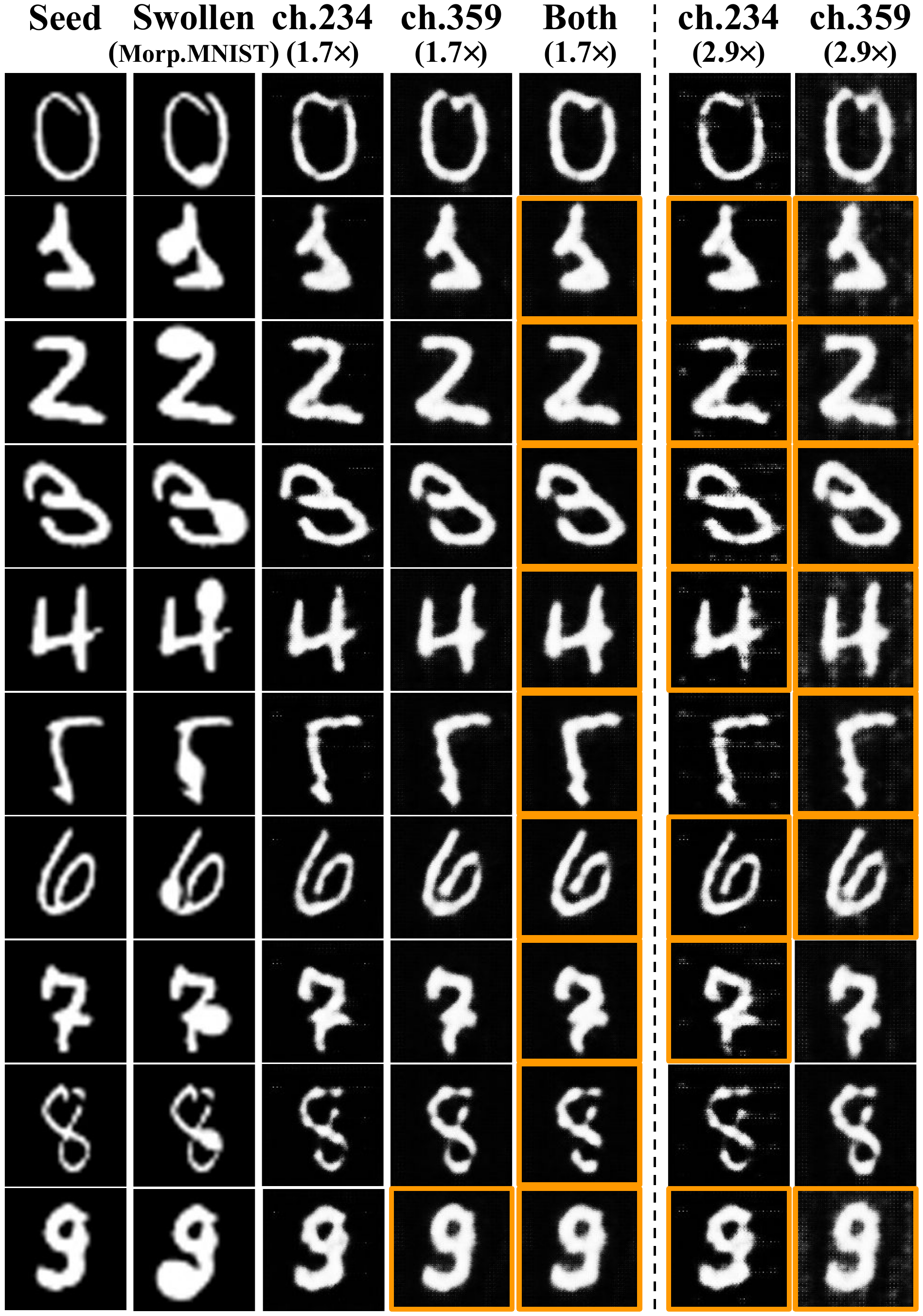}
    \captionof{figure}{(Extension of Fig.\,\ref{fig:multi-ch-test}) Testing the interactions of two channels in AlexNet-SW. It shows the images that are generated for both {\it ch.234} and {\it ch.359} channels as well as the images that are generated for each of the channels; the numbers in parentheses denote the channel intensities.}
    \label{fig:multi-ch-test-ext}
\end{figure}

\begin{figure*}[t!]
    \centering
    \includegraphics[width=0.80\textwidth]{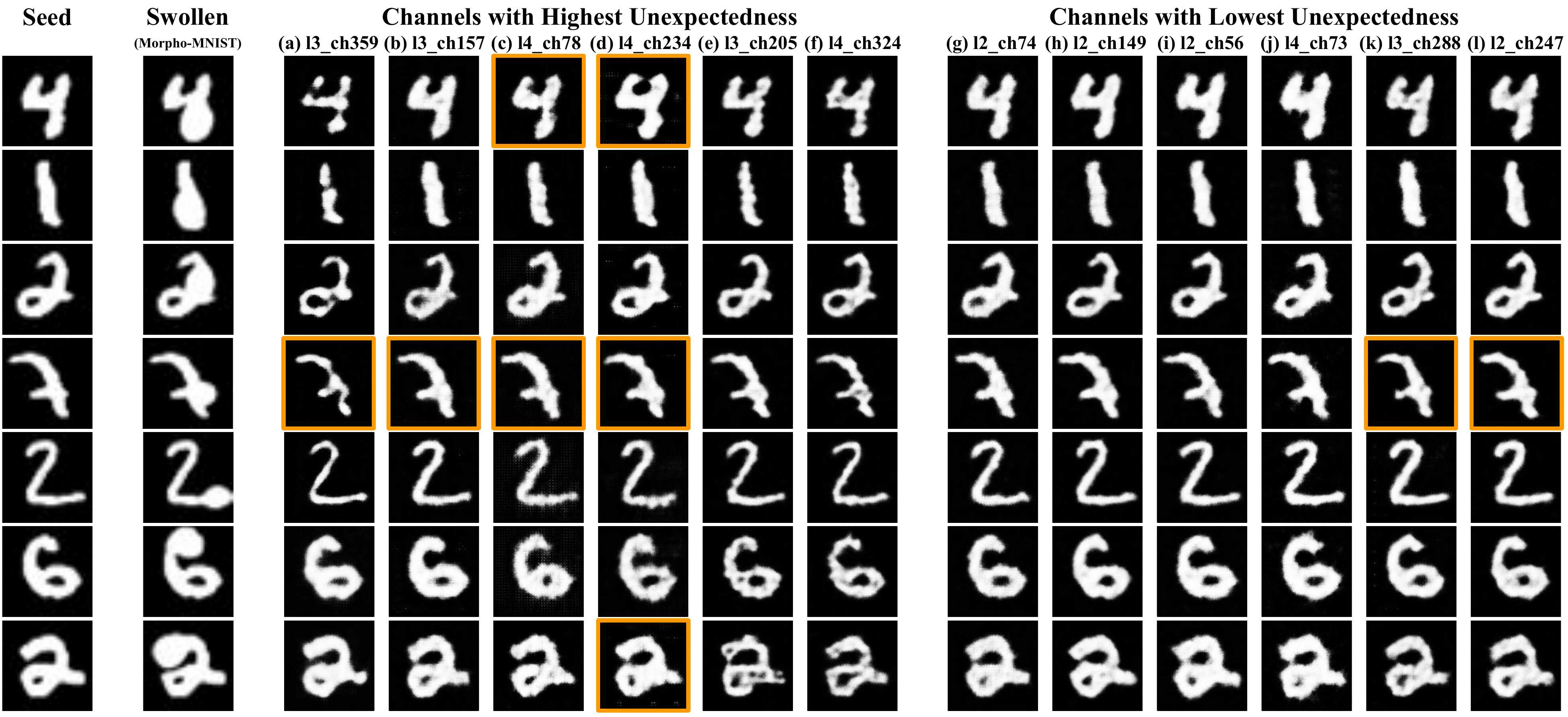}
    \caption{Example images generated for twenty selected channels of AlexNet-SW. 
    Two left-most columns show the seed images in MNIST and the corresponding swollen images in Morpho-MNIST. The next six columns are the channels that have highest unexpectedness scores (by the raw and normalized scores). The right-most six columns are the channels with lowest unexpectedness scores.}
    \label{fig:swell_img}
\end{figure*}

\begin{table}[t!]
    \scriptsize
    \centering
    \caption{Unexpectedness Scores for AlexNet-SW's 20 channels.}
    
    \begin{tabular}{|l|cr|cr|}
        \hline
        \multicolumn{1}{|c|}{\multirow{2}{*}{Channel}} & 
        \multicolumn{2}{c|}{Unexpectedness (Norm'ed)} & 
        \multicolumn{2}{c|}{Unexpectedness (Raw)} \\ \cline{2-5} 
        \multicolumn{1}{|c|}{} & \multicolumn{1}{c|}{Rank} & \multicolumn{1}{c|}{Score} & \multicolumn{1}{c|}{Rank} & \multicolumn{1}{c|}{Score} \\ \hline
        l3\_ch359 & \multicolumn{1}{c|}{1} & \textbf{0.0605} & \multicolumn{1}{c|}{3} & \textbf{98.757} \\ \hline
        l3\_ch157 & \multicolumn{1}{c|}{2} & \textbf{0.0517} & \multicolumn{1}{c|}{4} & \textbf{84.358} \\ \hline
        l4\_ch78 & \multicolumn{1}{c|}{3} & \textbf{0.0500} & \multicolumn{1}{c|}{1} & \textbf{183.642} \\ \hline
        {\ul l4\_ch234} & \multicolumn{1}{c|}{4} & \textbf{0.0275} & \multicolumn{1}{c|}{2} & \textbf{100.986} \\ \hline
        l3\_ch205 & \multicolumn{1}{c|}{5} & \textbf{0.0270} & \multicolumn{1}{c|}{9} & 44.095 \\ \hline
        l3\_ch64 & \multicolumn{1}{c|}{6} & 0.0267 & \multicolumn{1}{c|}{10} & 43.568 \\ \hline
        l2\_ch113 & \multicolumn{1}{c|}{7} & 0.0250 & \multicolumn{1}{c|}{11} & 40.801 \\ \hline
        l4\_ch324 & \multicolumn{1}{c|}{8} & 0.0212 & \multicolumn{1}{c|}{5} & \textbf{77.953} \\ \hline
        l4\_ch6 & \multicolumn{1}{c|}{9} & 0.0198 & \multicolumn{1}{c|}{6} & 72.872 \\ \hline
        l4\_ch339 & \multicolumn{1}{c|}{10} & 0.0184 & \multicolumn{1}{c|}{12} & 30.038 \\ \hline
        l2\_ch14 & \multicolumn{1}{c|}{11} & 0.0177 & \multicolumn{1}{c|}{13} & 28.967 \\ \hline
        l4\_ch254 & \multicolumn{1}{c|}{12} & 0.0146 & \multicolumn{1}{c|}{7} & 53.533 \\ \hline
        l3\_ch12 & \multicolumn{1}{c|}{13} & 0.0143 & \multicolumn{1}{c|}{15} & 23.409 \\ \hline
        l2\_ch247 & \multicolumn{1}{c|}{14} & 0.0124 & \multicolumn{1}{c|}{16} & 20.279 \\ \hline
        l4\_ch335 & \multicolumn{1}{c|}{15} & 0.0122 & \multicolumn{1}{c|}{8} & 44.906 \\ \hline
        l3\_ch288 & \multicolumn{1}{c|}{16} & 0.0076 & \multicolumn{1}{c|}{17} & 12.404 \\ \hline
        l4\_ch73 & \multicolumn{1}{c|}{17} & 0.0069 & \multicolumn{1}{c|}{14} & 25.322 \\ \hline
        l2\_ch56 & \multicolumn{1}{c|}{18} & 0.0046 & \multicolumn{1}{c|}{18} & 7.445 \\ \hline
        l2\_ch149 & \multicolumn{1}{c|}{19} & 0.0036 & \multicolumn{1}{c|}{19} & 5.797 \\ \hline
        l2\_ch74 & \multicolumn{1}{c|}{20} & 0.0034 & \multicolumn{1}{c|}{20} & 5.600 \\ \hline
    \end{tabular}
    \label{tbl:swell-unexpectedness}
\end{table}

\begin{table*}[t!]
    \scriptsize
    \centering
    \caption{(Extension of Table \ref{tbl:defect-stat}) The rate of generated test data for the defective channel
    that produce incorrect inference.
    The rankings of the defective channels by unexpectedness score are shown in parentheses.}
    
    \begin{tabular}{|c|ccc|l|c|c|l|c|cc|}
        \cline{1-4} \cline{6-7} \cline{9-11}
        \multirow{3}{*}{\textbf{\begin{tabular}[c]{@{}c@{}}Range of \\ Intensity $I$\end{tabular}}} & \multicolumn{3}{c|}{\textbf{Model (score rank / tested channels)}} &  & \multirow{3}{*}{\textbf{\begin{tabular}[c]{@{}c@{}}Range of \\ Intensity $I$\end{tabular}}} & \textbf{Model} &  & \multirow{3}{*}{\textbf{\begin{tabular}[c]{@{}c@{}}Range of \\ Intensity $I$\end{tabular}}} & \multicolumn{2}{c|}{\textbf{Model}} \\ \cline{2-4} \cline{7-7} \cline{10-11} 
         & \multicolumn{1}{c|}{\textbf{\begin{tabular}[c]{@{}c@{}}AlexNet-TH\\ (2/20)\end{tabular}}} & \multicolumn{1}{c|}{\textbf{\begin{tabular}[c]{@{}c@{}}AlexNet-SW\\ (4/20)\end{tabular}}} & \textbf{\begin{tabular}[c]{@{}c@{}}VGG-hair\\ (3/40)\end{tabular}} &  &  & \textbf{\begin{tabular}[c]{@{}c@{}}VGG-skin\\ (3/40)\end{tabular}} &  &  & \multicolumn{1}{c|}{\textbf{\begin{tabular}[c]{@{}c@{}}ResNet-hair\\ (2/60)\end{tabular}}} & \textbf{\begin{tabular}[c]{@{}c@{}}ResNet-skin\\ (1/60)\end{tabular}} \\ \cline{1-4} \cline{6-7} \cline{9-11} 
        0.9 -- 2.0 & \multicolumn{1}{c|}{31.8\%} & \multicolumn{1}{c|}{16.8\%} & 56.7\% &  & 1.0 -- 4.0 & 7.3\% &  & 1.0 -- 1.5 & \multicolumn{1}{c|}{19.29\%} & 17.58\% \\ \cline{1-4} \cline{6-7} \cline{9-11} 
        0.7 -- 3.0 & \multicolumn{1}{c|}{67.8\%} & \multicolumn{1}{c|}{56.2\%} & 80.4\% &  & 1.0 -- 5.0 & 24.7\% &  & 1.0 -- 2.5 & \multicolumn{1}{c|}{25.65\%} & 30.61\% \\ \cline{1-4} \cline{6-7} \cline{9-11} 
        0.5 -- 4.0 & \multicolumn{1}{c|}{84.8\%} & \multicolumn{1}{c|}{85.2\%} & 87.4\% &  & 1.0 -- 6.0 & 59.9\% &  & 1.0 -- 4.0 & \multicolumn{1}{c|}{33.73\%} & 44.85\% \\ \cline{1-4} \cline{6-7} \cline{9-11} 
        0.4 -- 4.5 & \multicolumn{1}{c|}{89.0\%} & \multicolumn{1}{c|}{90.2\%} & 89.4\% &  & 1.0 -- 8.0 & 88.8\% &  & 1.0 -- 6.0 & \multicolumn{1}{c|}{43.93\%} & 56.77\% \\ \cline{1-4} \cline{6-7} \cline{9-11}
    \end{tabular}
    \label{tbl:defect-stat-all}
\end{table*}

Fig.\,\ref{fig:swell_img} shows some of the test images generated with the twelve channels, i.e., six channels with the highest unexpectedness (column a--f) and six with the lowest unexpectedness (column g--l). 
It also shows the seed images in MNIST and their corresponding swollen images in Morph-MNIST.
We observed that the channels with high unexpectedness 
show swelling strokes in some of the generated images, although the swelling is most notable in channel l4\_ch234 (column d). 
We also observed that seven test images out of forty-two (17\%) caused the change of inference outcome (marked by the orange boxes).
In comparison, the channels with low unexpectedness (column g--l) do not have noticeable variations.
Moreover, only two images (5\%) caused the change of inference outcome, which is for the digit seven
that is relatively close to the decision boundary.
After we look into other test images and their inference outcome,
we concluded that channel l4\_ch234 (column d)
is correlated with the stroke swelling, which causes the bias (towards zero) in the classification.

We also present the detailed results of testing the interactions of two channels by feeding the output of the {\FGAN} that is trained for 
one channel as the input of another {\FGAN} hat is trained for a different channel.
Fig.\,\ref{fig:multi-ch-test-ext} shows the results; the figure compares the images
that are generated for channel {\it ch.234} and {\it ch.359} separately (denoted by the channel names) as well as the images that are generated for both the channels 
(denoted by {\it Both}).
We can see that the images that are generated for both the channels cause 
more mis-classifications than the images that are generated for single channels 
(the orange boxes denote that the image sample is mis-classified by the tested CNN).
If we increase the intensity and generate images for each channel separately
the generated images tend to have more noises as shown at the right side of the figure.
We also observed that the images generated for both the channels have the 
attributes (or transformations) of the images that are generated with each of the channels.
Although further studies are required, we confirmed that we can test multiple CNN channels by chaining multiple {\FGAN} instances.

Table \ref{tbl:defect-stat-all} shows the rate of the generated test data for thick and
swollen channels resulting in mis-classification for varying intensity input I. We extended Table~\ref{tbl:defect-stat} in the manuscript with the results for the VGG-skin model, ResNet-hair and ResNet-skin.

\begin{figure}[th!]
    \centering
    \includegraphics[width=1.0\linewidth]{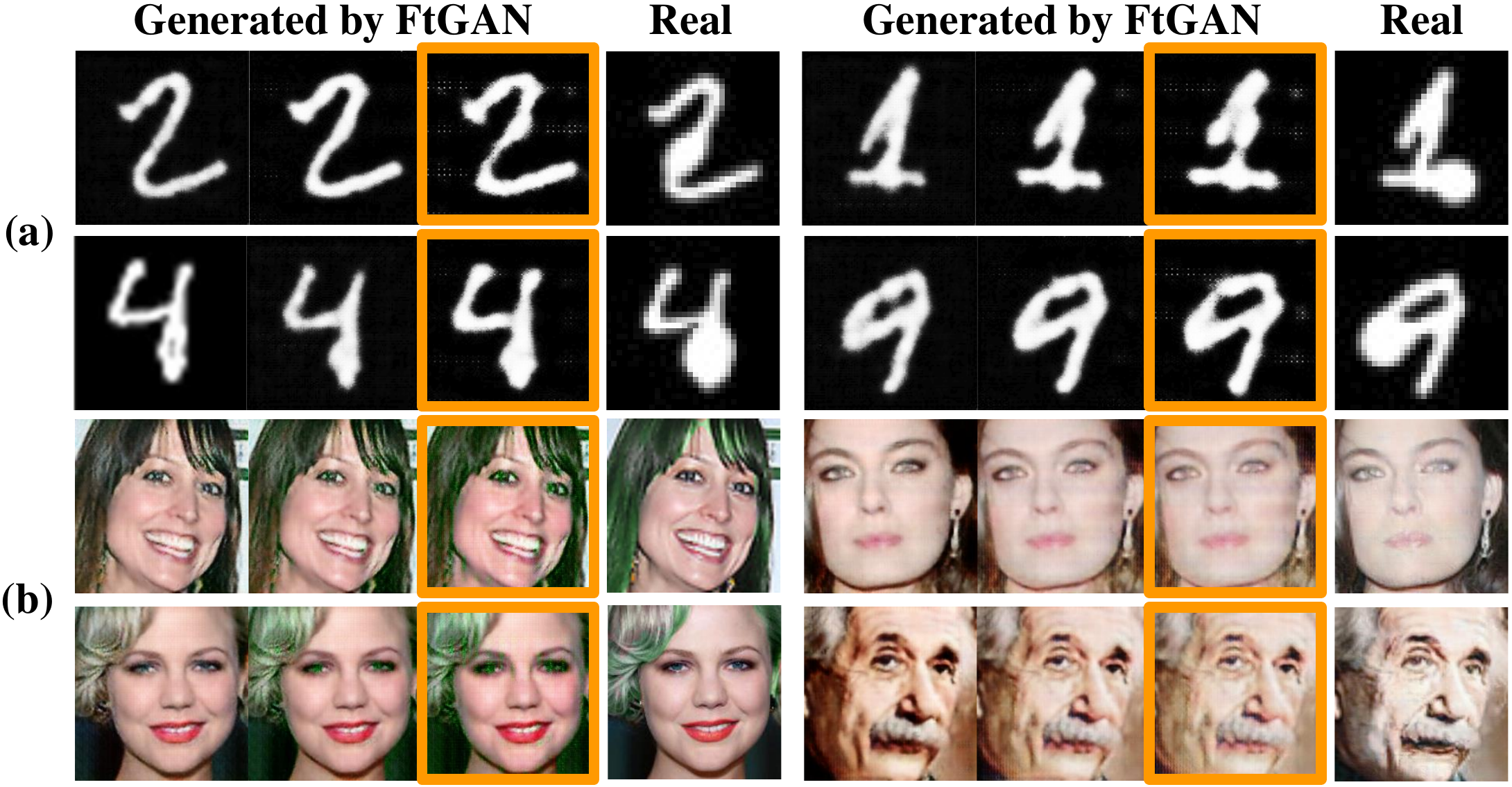}
    \caption{(Extension of Fig.\,\ref{fig:debug-result}) Example images generated for defect-injected CNN instances --AlexNet-swell ((a) left and right), VGG-hair ((b) left) and VGG-skin ((b) right).}
    \label{fig:ext_fig5}
\end{figure}

\begin{figure}[th!]
    \centering
    \includegraphics[width=1.0\linewidth]{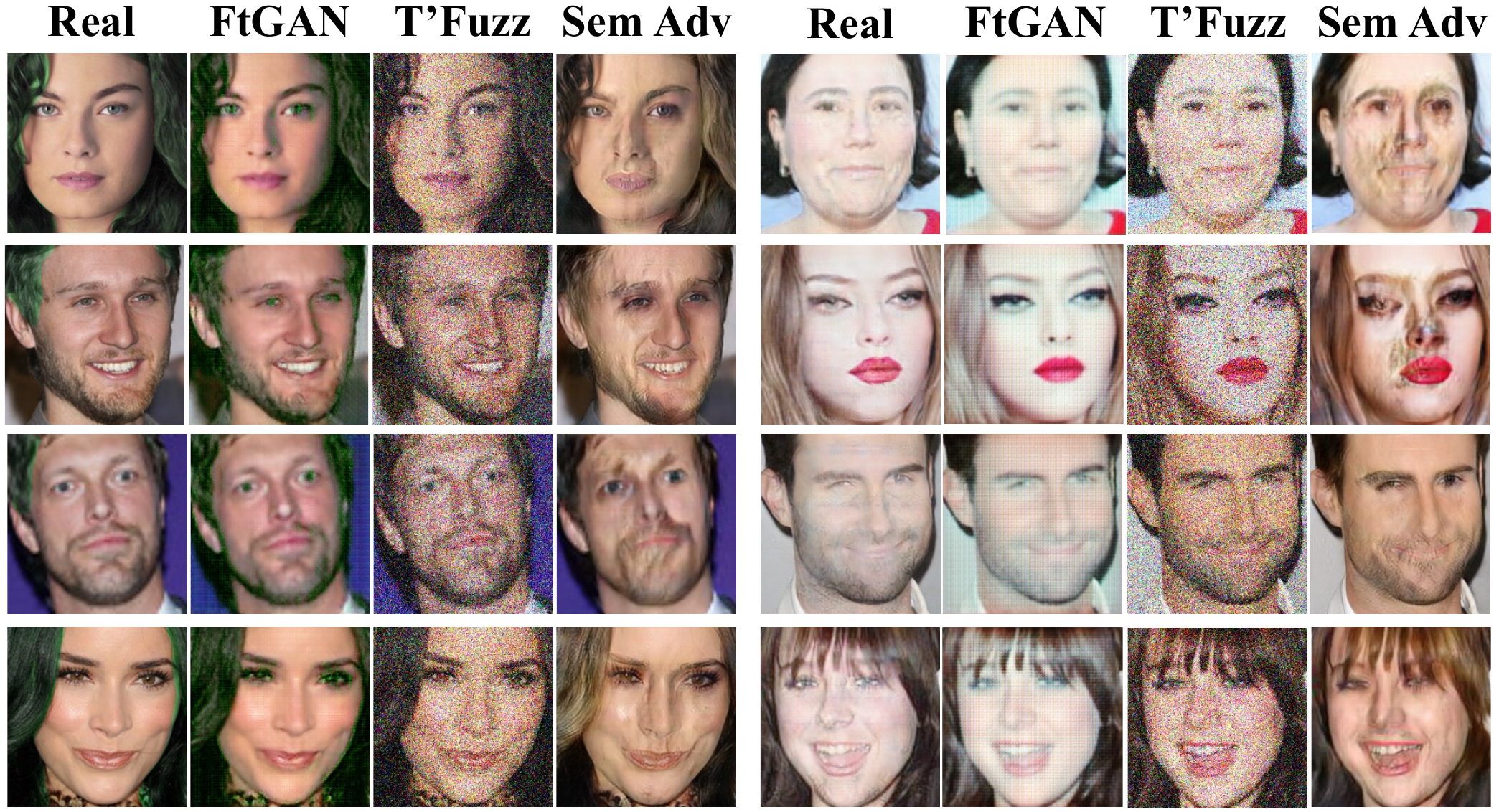}
    \caption{(Extension of Fig.\,\ref{fig:debug-cmp}) Example images generated by {\FGAN}, 
    TensorFuzz (T'Fuzz), and semantic adversarial attack (Sem Adv) for VGG-hair (left) and ResNet-skin (right).}
    \label{fig:ext_fig6}
\end{figure}

\noindent
{\bf Results for the Data Corruption Bug Detection.}
Fig.\,\ref{fig:ext_fig5} shows the example images generated for the defect-injected CNN instances in Section~\ref{sec:eval-find-problem}; (a) shows the images that are generated for AlexNet-SW (with varying channel intensities) and (b) shows the images that are generated for the VGG-hair and VGG-skin models.
The images in orange boxes are the defect-inducing test images. 
Fig.\,\ref{fig:ext_fig6} extends Fig.\,\ref{fig:debug-cmp} and compares the test images that are generated by {\FGAN}, TensorFuzz, 
and semantic adversarial attack.

\begin{figure}[h!]
\centering
\includegraphics[width=\linewidth]{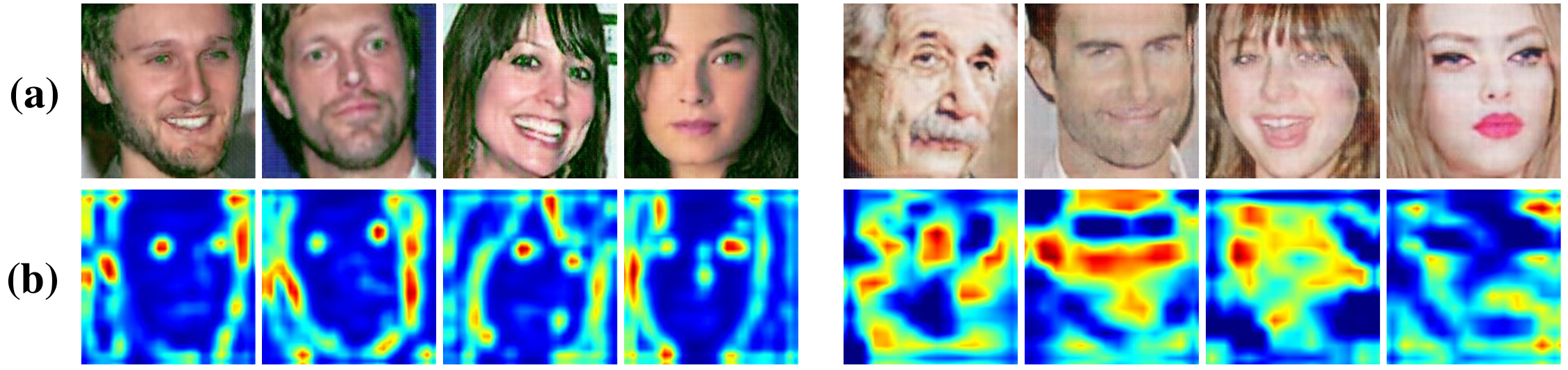}
        \caption{The results of applying Grad-CAM to \FGAN's test images for VGG-hair (left four) and VGG-skin (right four).}
        \label{fig:grad_cam}
\end{figure}

\noindent
{\bf Analysis with an XAI Technique.}
Furthermore, we applied Grad-CAM, a state-of-the-art XAI (eXplainable AI) technique, to the test images 
that are generated for the defective CNN instances (VGG-hair and VGG-skin).
The results are shown in Fig.\,\ref{fig:grad_cam}; (a) shows the test images (left four images are for VGG-hair
and right four images are for VGG-skin) and (b) shows the corresponding Grad-CAM results.
All the test images shown in the figure are defect-inducing ones (i.e. they are identified as 
the incorrect target identity that the models are trained for).
For the test images of VGG-hair, we observed that the green eyes and green hair colors
account for the (incorrect) inference computation. 
For the test images of VGG-skin, we can see that a large portion of the faces
account for the inference computation.
We can see that {\FGAN} successfully made the defect-inducing variations to the test images
and identified the defects in the two CNN instances.

\subsection{Finding Bugs in a Real CNN (Extension of Section\,\ref{sec:eval-public-cnn})}
\label{sec:exteval3}

\begin{figure}[t!]
    \centering
    \includegraphics[width=0.95\linewidth]{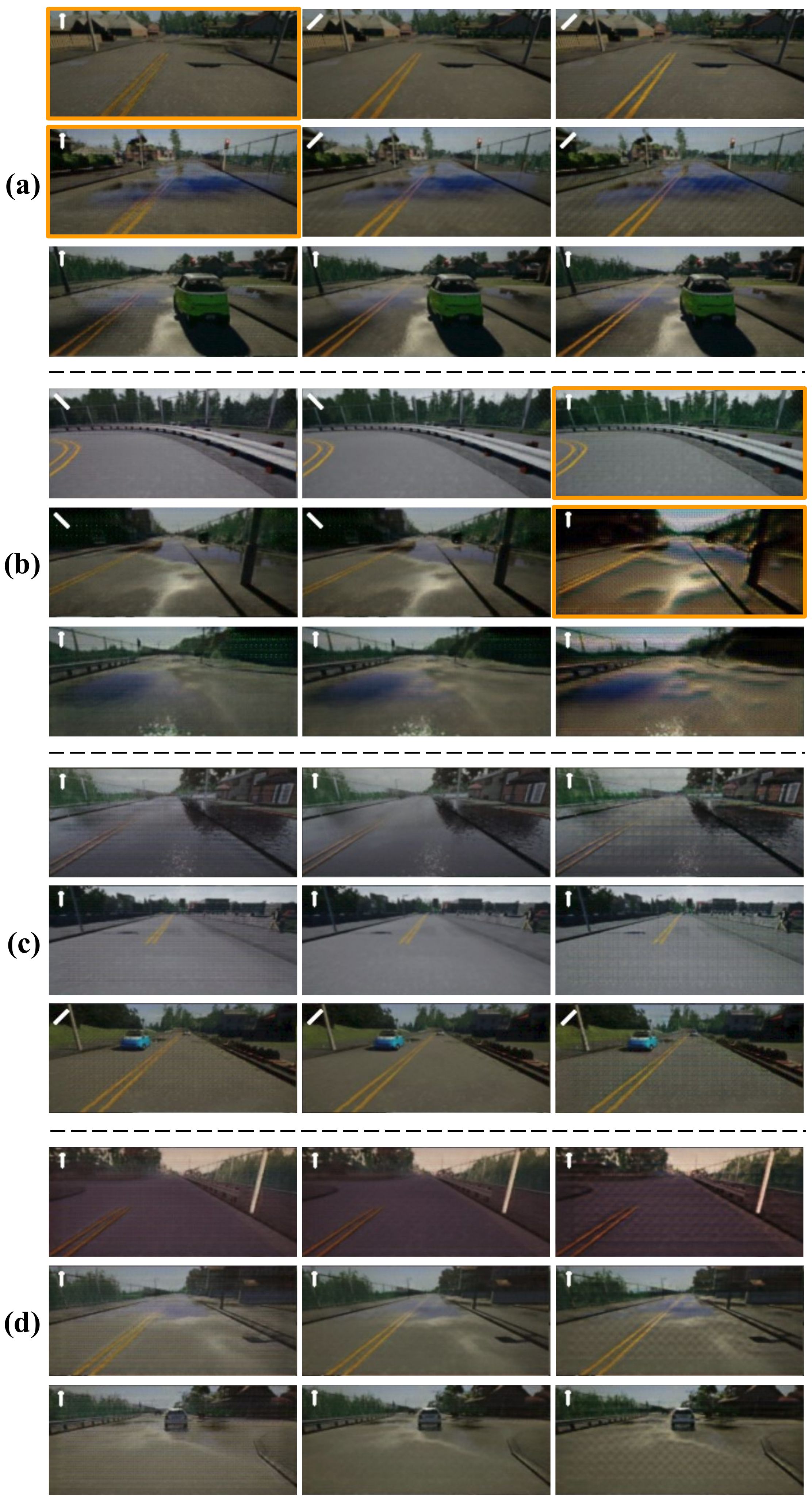}
    \caption{(Extension of Fig.\,\ref{fig:debug-carla}) Test images for CARLA-CNN; (a) is center line wear-out and (b) is road texture and color tone changes;
    (c) and (d) have non human-recognizable variations.
    }
    \label{fig:ext_fig8}
\end{figure}

Fig.\,\ref{fig:ext_fig8} shows {\FGAN}'s generated images for a publicly-available real-world CNN instance for autonomous driving. Again the images in orange boxes cause the CNN instance make incorrect steering decision. 
The images in (a) have the center line wear-out and the images in (b) have road-texture changes and the color tone changes.
The images in (c) and (d) do not seem to have human-recognizable semantic attributes;
they seem to add grid-pattern (or x-shape pattern) noises to the images.

\end{document}